\documentclass{article}
\usepackage{iclr2027_conference,times}

\usepackage{amsmath,amsfonts,bm}

\def\eqref#1{equation~\ref{#1}}

\def\1{\bm{1}}

\DeclareMathAlphabet{\mathsfit}{\encodingdefault}{\sfdefault}{m}{sl}
\SetMathAlphabet{\mathsfit}{bold}{\encodingdefault}{\sfdefault}{bx}{n}

\usepackage{booktabs}
\usepackage{array}
\usepackage{caption}
\usepackage{amssymb}
\usepackage{float}
\usepackage{graphicx}
\usepackage{hyperref}
\usepackage{url}
\usepackage{multirow}
\usepackage[section]{placeins}
\usepackage{capt-of}
\usepackage{enumitem}
\newcommand{\CheckmarkBold}{\(\bm{\checkmark}\)}
\iclrfinalcopy
\author{\small Jinlin Gan$^{1\ast}$ \quad Mingzhe Li$^{1\ast}$ \quad Junsheng Zhang$^{1}$ \quad Zhibo Zhou$^{1}$ \quad Jiahan Geng$^{1}$ \\
\small Yuhua Luo$^{1}$ \quad Changwang Zhang$^{2}$ \quad Chenglu Wen$^{1}$ \quad Lan Xu$^{3}$ \quad Siqi Shen$^{1\dagger}$ \quad Cheng Wang$^{1}$ \\
\small $^{1}$Xiamen University \quad $^{2}$Chery NEXTAI Research Institute \quad $^{3}$ShanghaiTech University}
\title{HumanoidPose: Monocular 3D Pose Estimation for Humanoid Robots}

\begin{document}

\maketitle
\footnotetext{$^\ast$ Equal contribution.}
\footnotetext{$^\dagger$ Corresponding author.}

\vspace{-3em}
\begin{figure}[H]
    \centering
    \includegraphics[width=\linewidth]{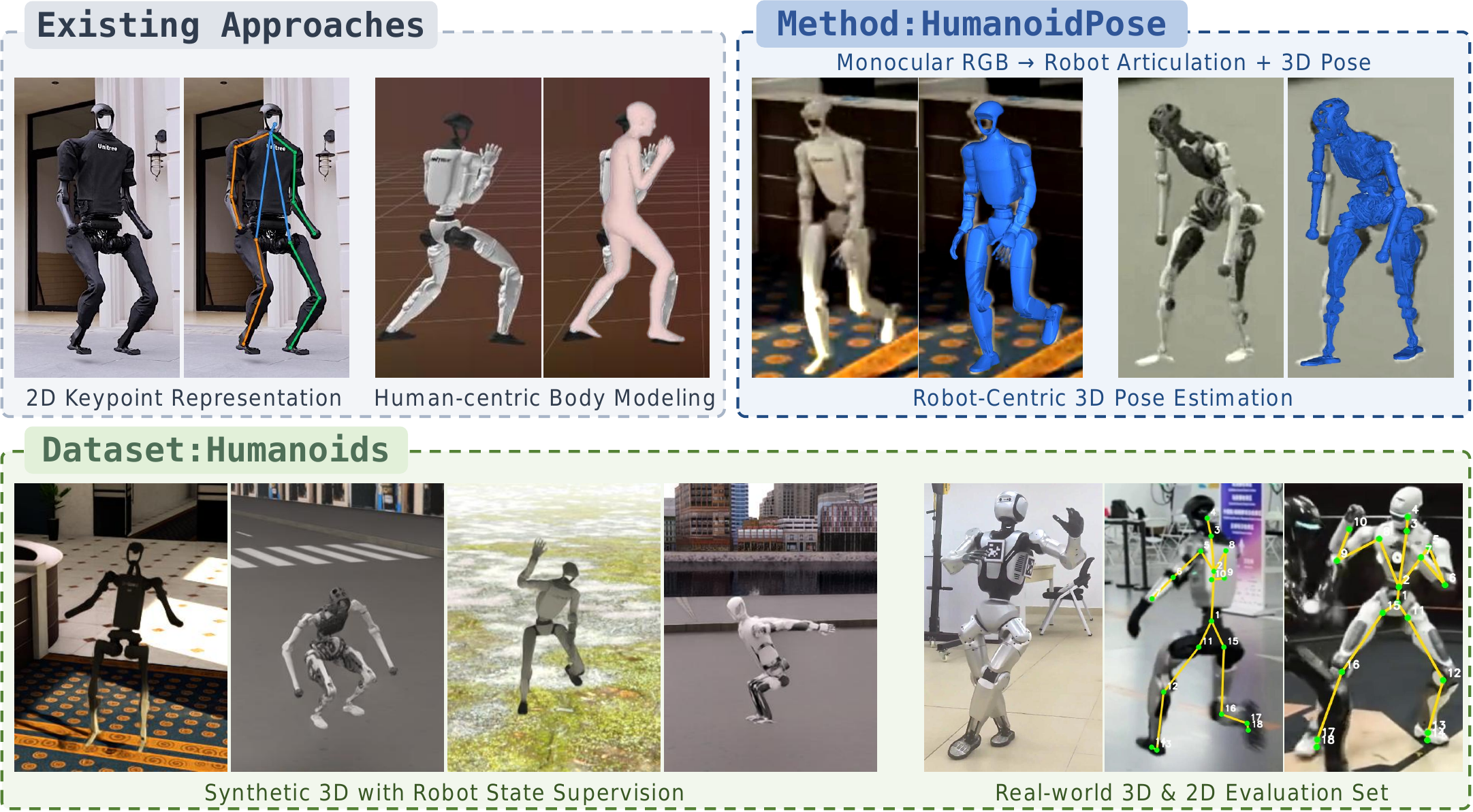}

    \caption{\textbf{Overview.} \textbf{To address the gap in monocular 3D humanoid robot pose estimation (HRPE)}, we introduce \textbf{Humanoids, a dedicated dataset and benchmark} spanning 13 humanoid embodiments, with a synthetic split for 3D supervision and a real-world split for evaluation, and \textbf{HumanoidPose, a dedicated framework} for monocular 3D humanoid robot pose estimation.}
    \label{fig:overview}
\end{figure}
\vspace{-0.8em}

\begin{abstract}

As humanoid robots increasingly enter real-world environments, recovering their 3D whole-body pose from monocular imagery is important for external state perception, motion analysis, and robot learning.
However, Humanoid Robot Pose Estimation (\textbf{HRPE}) lies at the intersection of
conventional Robot Pose Estimation (RPE) and Human Pose Estimation (HPE),
while not being fully addressed by either paradigm: RPE primarily focuses on non-humanoid robots, particularly fixed-base manipulators, whereas HPE is built upon human-specific representations. Despite recent progress in humanoid 2D keypoint detection and HPE adaptation, humanoid-native 3D pose recovery remains insufficiently studied, and a dedicated benchmark with suitable training data and evaluation protocols is still lacking.
To address this gap, we introduce \textbf{Humanoids}, a benchmark for
monocular 3D humanoid robot pose estimation, and propose
\textbf{HumanoidPose}, a framework for this task.
Humanoids provides synthetic 3D supervision across diverse humanoid embodiments,
manually annotated real-world 2D data, and real-world RGB sequences paired with
robot states for 3D evaluation on a physical Roban humanoid. HumanoidPose
combines \textbf{human-inspired visual perception} with robot
\textbf{articulation modeling} and explicit \textbf{kinematic grounding}.
Experiments on \emph{Humanoids} and the \emph{DHRP-H1} benchmark demonstrate
strong performance in articulation recovery and 3D pose estimation, including
zero-shot transfer to real-world 2D and 3D evaluation.
Ablation studies further show the contribution of robot-native articulation
modeling and explicit kinematic grounding.
The dataset and code will be publicly released to support future research in the community.

\end{abstract}

\section{Introduction}

Humanoid robots are increasingly entering real-world environments. Recovering their 3D whole-body states from external visual observations can support multi-robot collaboration and robot learning, including humanoid video-to-motion and sim-to-real-to-sim feedback. However, monocular 3D humanoid robot pose estimation remains largely unexplored.

Humanoid Robot Pose Estimation (\textbf{HRPE}) lies at the intersection of conventional
Robot Pose Estimation (RPE) and Human Pose Estimation (HPE), while not being fully addressed by either paradigm. RPE methods primarily focus on non-humanoid robots, particularly fixed-base manipulators, and do not readily handle the floating-base, freely moving, high-DoF whole-body motion of humanoids~\citep{labbe2021single,ban2024real,tian2024robokeygen,goswami2025robopepp}.
HPE methods are closer in motion structure, but are built upon human-specific representations that do
not directly correspond to robot articulation~\citep{Kanazawa2018EndtoEndRO,goel2023humans,SMPL2015}.

Existing humanoid-specific efforts mainly focus on 2D keypoint detection,
specialized sensing setups, or the direct application of existing HPE methods to
humanoid robots~\citep{heo2024diverse,cho2024full,young2025robots,li2026towards},
leaving monocular recovery of robot articulation and root pose insufficiently studied.
As summarized in Table~\ref{tab:data_compare}, progress is further limited
by the lack of datasets and benchmarks that jointly provide monocular RGB, robot joint states,
root pose, and accurate 3D supervision.

To address these gaps, we introduce \textbf{Humanoids, a dedicated dataset and benchmark} for monocular 3D humanoid robot pose estimation, and propose
\textbf{HumanoidPose, a task-specific framework} for this task.

\emph{Humanoids} covers \textbf{13} humanoid embodiments, including Unitree G1,
PND AdamLite, and Roban. For 12 embodiments, it
contains 3,706 source motions and 6,885 synthetic clips (\textbf{732K RGB frames})
with synchronized robot joint states, root poses, and FK-derived 3D keypoints,
as well as \textbf{365 real-world clips} with manually verified 2D annotations.
For Roban, we provide synthetic data and real-world RGB sequences paired with
joint states and estimated camera-relative root poses for 3D evaluation.
Synthetic annotations are verified through forward-kinematics and reprojection
\textbf{consistency checks}, while the original real-world 2D annotations undergo
temporal and kinematic \textbf{quality control}. A canonical 2D keypoint
representation provides a unified visual interface across humanoid embodiments.

\emph{HumanoidPose} directly estimates robot \textbf{articulation} and root pose from monocular RGB, \textbf{combining a human-inspired visual frontend and robot-native modeling.} We fuse target-centric visual features and human-inspired 2D keypoints through keypoint-conditioned joint queries for articulation prediction, while visual and bounding-box cues guide root-pose initialization. The predicted state is further refined using explicit robot kinematics and reprojection.

Experiments on \emph{Humanoids} and the external \emph{DHRP-H1} benchmark
demonstrate strong performance in robot articulation recovery and 3D pose
estimation. On real-world data, we evaluate zero-shot transfer through 2D pose
estimation and 3D state recovery on a physical Roban humanoid.

Our main contributions are summarized as follows:
\begin{itemize}[leftmargin=1.2em,labelsep=0.4em,topsep=2pt,partopsep=0pt,itemsep=2pt,parsep=0pt]
    \item \textbf{Early Systematic Investigation of HRPE.} We formulate monocular 3D Humanoid Robot Pose Estimation (HRPE) as a distinct visual perception task and conduct an early systematic investigation of this problem, differing from conventional Robot Pose Estimation (RPE) and Human Pose Estimation (HPE).
    \item \textbf{Dedicated Benchmark and Method.} We introduce \textbf{Humanoids}, a dataset and benchmark providing synthetic 3D supervision and real-world 2D and 3D evaluation across diverse humanoid embodiments, motions, and visual conditions. We further propose \textbf{HumanoidPose}, a task-specific framework combining human-inspired visual perception with robot-native articulation modeling and explicit kinematic grounding.
    \item \textbf{Comprehensive Evaluation.} We evaluate representative RPE and HPE approaches under a unified HRPE protocol across synthetic and real-world settings. Extensive experiments and controlled ablations demonstrate the effectiveness of HumanoidPose and provide insights into the challenges and effective designs for monocular 3D HRPE.
\end{itemize}

\section{Related Work}

\subsection{Robot Pose Estimation}

Robot Pose Estimation (RPE) aims to recover robot articulation and spatial
pose from visual observations. Existing methods explore render-and-compare
optimization, keypoint-based 3D lifting, holistic feed-forward pose estimation,
and self-supervised visual representation learning~\citep{labbe2021single,tian2024robokeygen,ban2024real,goswami2025robopepp,liu2026robotagendtoendrobotconfiguration}.
However, existing RPE methods primarily target non-humanoid robots,
particularly fixed-base manipulators, and do not readily extend to recovering
robot articulation and root pose for floating-base humanoids undergoing
high-DoF whole-body motion.

Humanoid 2D keypoint estimation has been studied~\citep{heo2024diverse,young2025robots},
while full-body pose estimation with head-worn cameras for digital
human-augmented robotic telepresence has been investigated~\citep{cho2024full}.
However, these approaches focus on 2D landmarks or specialized sensing,
rather than monocular recovery of 3D robot articulation states and
camera-relative global pose.

\subsection{Human Pose and Motion Recovery}

Human Pose Estimation (HPE) and Human Mesh Recovery (HMR) are closer to
humanoid pose estimation in whole-body motion structure. Methods built on
human-body models such as SMPL~\citep{SMPL2015} include
optimization-based fitting such as SMPLify~\citep{Bogo2016KeepIS},
video-based recovery such as VIBE~\citep{VIBE}, and world-grounded methods
such as WHAM~\citep{shin2024wham} and GVHMR~\citep{shen2024world}, which
support monocular 3D body and motion recovery. However, their human-specific
skeletal topologies, joint definitions, body models, and motion priors do not
directly correspond to robot articulation.

The Motion Turing Test (MTT)~\citep{li2026towards} and
I-Robot~\citep{kim2026robot} represent humanoid motions through human-body
pose models for motion analysis and evaluation. However, monocular recovery of
robot articulation and root pose remains insufficiently studied. HumanoidPose
instead directly recovers these robot-native states from monocular RGB
observations.

\subsection{Pose and Motion Datasets for Humans and Robots}

Human datasets such as Human3.6M~\citep{Ionescu14a},
AMASS~\citep{AMASS_ICCV2019}, Motion-X~\citep{lin2023motion}, and
LAFAN1~\citep{harvey2020robust} provide large-scale pose and motion
supervision, while robot datasets such as DREAM~\citep{lee2020camera} and
RoboNet~\citep{dasari2019robonet} mainly focus on non-humanoid embodiments.
Recent humanoid datasets, including DHRP~\citep{heo2024diverse},
HumanoidRobotPose~\citep{amini2021real}, PHUMA~\citep{lee2025phuma}, and
HHMotion~\citep{li2026towards}, further advance humanoid robot research from perception, motion,
and control perspectives.

As summarized in Table~\ref{tab:data_compare}, existing datasets remain
limited in jointly providing visual observations, 3D pose annotations, and
robot-state supervision across diverse humanoid embodiments.
\textbf{Humanoids} is designed to address this gap.

\begin{figure}[h]
    \centering

    \includegraphics[width=\linewidth]{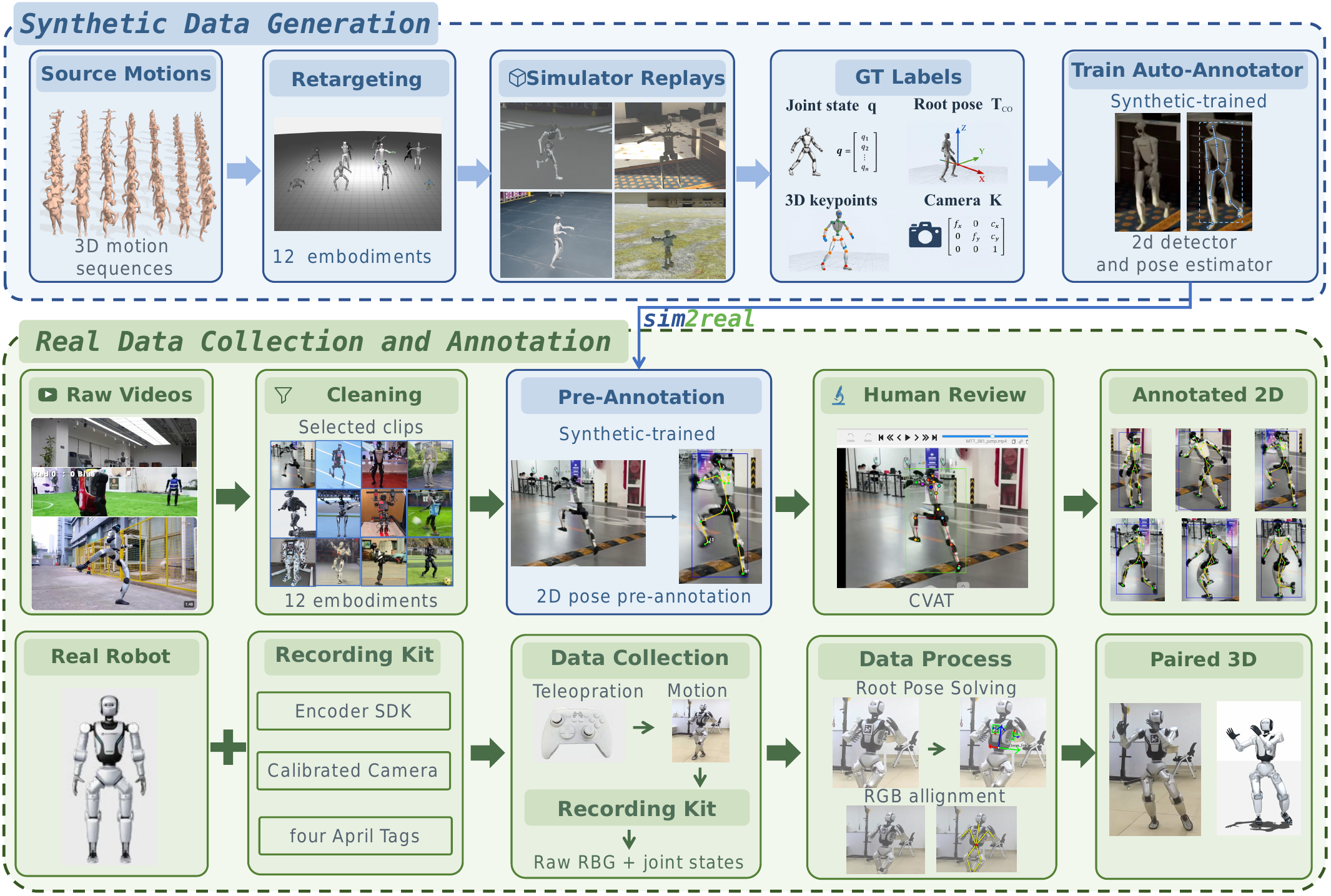}
    \caption{\textbf{\emph{Humanoids} Dataset Construction Pipeline}. From left to right, the original dataset pipeline illustrates synthetic 3D data generation and real-world 2D annotation, where synthetic-trained 2D pose estimator enables sim-to-real pre-annotation.
    }
    \label{fig:dataset_overview}

    \vspace{1mm}

\end{figure}
\begin{table}[h]
\centering
\caption{
Comparison with related datasets.
The ``-'' symbol indicates that the corresponding data type or annotation is unavailable.}

\label{tab:data_compare}

\vspace{1mm}

{\scriptsize

\renewcommand{\arraystretch}{0.85}
\setlength{\tabcolsep}{2pt}

\resizebox{\linewidth}{!}{
\begin{tabular}{lccccccccc}

\toprule[1pt]

\multirow{2}{*}{Dataset}
& \multirow{2}{*}{Domain}
& \multicolumn{4}{c}{Data Composition}
& \multirow{2}{*}{Real/Sim}
& \multirow{2}{*}{\#Emb.}
& \multirow{2}{*}{\#Dur}
& \multirow{2}{*}{\#Clips}
\\

\cmidrule{3-6}

&
& RGB
& 2D KP
& 3D Pose
& Robot State
&
&
&
&
\\

\midrule

AMASS\citep{AMASS_ICCV2019}
& Human
& -
& -
& \CheckmarkBold
& -
& Real
& -
& 40+h
& 26.3k
\\

Human3.6M\citep{Ionescu14a}
& Human
& \CheckmarkBold
& \CheckmarkBold
& \CheckmarkBold
& -
& Real
& -
& -
& -
\\

Motion-X\citep{lin2023motion}
& Human
& \CheckmarkBold
& -
& \CheckmarkBold
& -
& Real
& -
& 144.2h
& 81,084
\\

LAFAN1\citep{harvey2020robust}
& Human
& -
& -
& \CheckmarkBold
& -
& Real
& -
& 4.6h
& -
\\

DREAM\citep{lee2020camera}
& Non-Humanoid
& \CheckmarkBold
& \CheckmarkBold
& \CheckmarkBold
& \CheckmarkBold
& Real+Sim
& 3
& -
& -
\\

RoboNet\citep{dasari2019robonet}
& Non-Humanoid
& \CheckmarkBold
& -
& -
& \CheckmarkBold
& Real
& 7
& -
& {162,417}
\\

DHRP\citep{heo2024diverse}
& Humanoid
& \CheckmarkBold
& \CheckmarkBold
& -
& -
& Real+Sim
& 11
& -
& 462
\\

HumanoidRobotPose\citep{amini2021real}
& Humanoid
& \CheckmarkBold
& \CheckmarkBold
& -
& -
& Real
& 10+
& -
& 23
\\

BONES-SEED\citep{luo2026sonic}
& Human-Humanoid
& -
& -
& \CheckmarkBold
& \CheckmarkBold
& Real
& -
& -
& -
\\

PHUMA\citep{lee2025phuma}
& Human-Humanoid
& -
& -
& \CheckmarkBold
& \CheckmarkBold
& Sim
& 2
& 72.96h
& 76.01k
\\

HHMotion\citep{li2026towards}
& Human-Humanoid
& \CheckmarkBold
& -
& \CheckmarkBold
& -
& Real+Sim
& 11
& 1.39h
& 1,000
\\

\midrule

\textcolor[rgb]{0.066,0.466,0.69}{\textbf{Humanoids (Ours)}}
& \textcolor[rgb]{0.066,0.466,0.69}{Humanoid}
& \textcolor[rgb]{0.066,0.466,0.69}{\CheckmarkBold}
& \textcolor[rgb]{0.066,0.466,0.69}{\CheckmarkBold}
& \textcolor[rgb]{0.066,0.466,0.69}{\CheckmarkBold}
& \textcolor[rgb]{0.066,0.466,0.69}{\CheckmarkBold}
& \textcolor[rgb]{0.066,0.466,0.69}{Real+Sim}
& \textcolor[rgb]{0.066,0.466,0.69}{\textbf{13}}
& \textcolor[rgb]{0.066,0.466,0.69}{\textbf{8.1h}}
& \textcolor[rgb]{0.066,0.466,0.69}{\textbf{7,848}}
\\

\bottomrule[1pt]

\end{tabular}
}

}

\vspace{-3mm}

\end{table}

\section{Humanoids: Dataset and Benchmark}
\subsection{Overview}

Existing datasets provide limited support for humanoid robot 3D pose estimation
(Table~\ref{tab:data_compare}).

We therefore introduce \textbf{Humanoids}, a dataset and benchmark for
monocular 3D humanoid robot pose estimation with RGB observations, robot joint states,
root poses, and FK-derived 3D keypoints.
The original synthetic set covers 12 humanoid embodiments, 3,706 unique source motions, 6,885 motion clips,
and 732K RGB frames across diverse motions and visual conditions.
In addition, we provide a real-world evaluation set containing 365 RGB clips,
56,437 frames, and 12 embodiments, with manually verified 2D annotations using
a Canonical-18 keypoint representation (Appendix~\ref{app:canonical18}) for
evaluating real-world generalization.
We extend Humanoids with a thirteenth embodiment, Roban: 574 synthetic clips
(60,232 frames) and 24 real-world sequences (23,170 frames) with SDK joint
states and camera-relative root poses estimated using AprilTags. This extension
enables real-world 3D evaluation (Appendix~\ref{app:roban}).

\subsection{Data Construction}
\subsubsection{Synthetic data}

Human motions from the HDM05, CMU, and BMLrub subsets of AMASS \citep{AMASS_ICCV2019} are retargeted
to 12 diverse humanoid embodiments (listed in Appendix~\ref{sec:perrobot}) using GMR~\citep{araujo2025retargeting} and replayed in Isaac Sim \citep{isaac_sim} and MuJoCo \citep{todorov2012mujoco}.
Robot joint states, root poses, and FK-derived Canonical-18 3D keypoints are
recorded directly from simulation to provide synchronized robot-state and 3D pose supervision.
The detailed definition of the Canonical-18 keypoint representation is provided
in Appendix~\ref{app:canonical18}.
We include diverse simulated environments and randomize robot placements,
camera viewpoints, distances, and lighting conditions, and render
$640\times480$ RGB frames with synchronized camera parameters.
After clip extraction and quality control, the original synthetic set contains
6,885 clips and 732K frames. We additionally generate 574 synthetic Roban
clips (60,232 frames), split into 450 training, 55 validation, and 69 test clips.

\subsubsection{Real-world data}

We collect public and manufacturer-provided humanoid videos covering the same
12 robot embodiments as the original synthetic set, with diverse viewpoints,
environments, and motions. We remove sequences with severe occlusions or poor
visibility, and segment videos into 2--8s clips containing individual
continuous motions.

For annotation, we leverage an off-the-shelf detector~\citep{redmon2016you} and
keypoint model~\citep{xu2022vitpose} trained on the synthetic benchmark to
initialize robot bounding boxes and Canonical-18 keypoint predictions on
real-world videos. The annotations are then manually
verified and corrected in CVAT~\citep{CVAT}. The resulting bounding boxes and
Canonical-18 keypoints support image-space evaluation under real-world
conditions.
For real-world 3D evaluation, we additionally capture a physical Roban humanoid
performing teleoperated motions with a calibrated external RGB camera. The
robot SDK records joint states, while four AprilTags mounted on the front,
back, left, and right of the torso, together with calibrated tag-to-root
transformations, provide estimates of the camera-relative root pose. The
resulting 24 sequences comprise 23,170 frames and include arm movements,
dances, walking, squatting, and combinations of these motions. These paired
RGB observations and robot states support real-world 3D evaluation; further
collection details are in Appendix~\ref{app:roban}.

\subsection{Dataset Quality}

\paragraph{Synthetic data.}

\begin{figure}[t]
\centering
\captionsetup{font=small,skip=5pt}
\begin{minipage}[t]{0.46\linewidth}
\vspace{0pt}
\centering

\includegraphics[width=\linewidth]{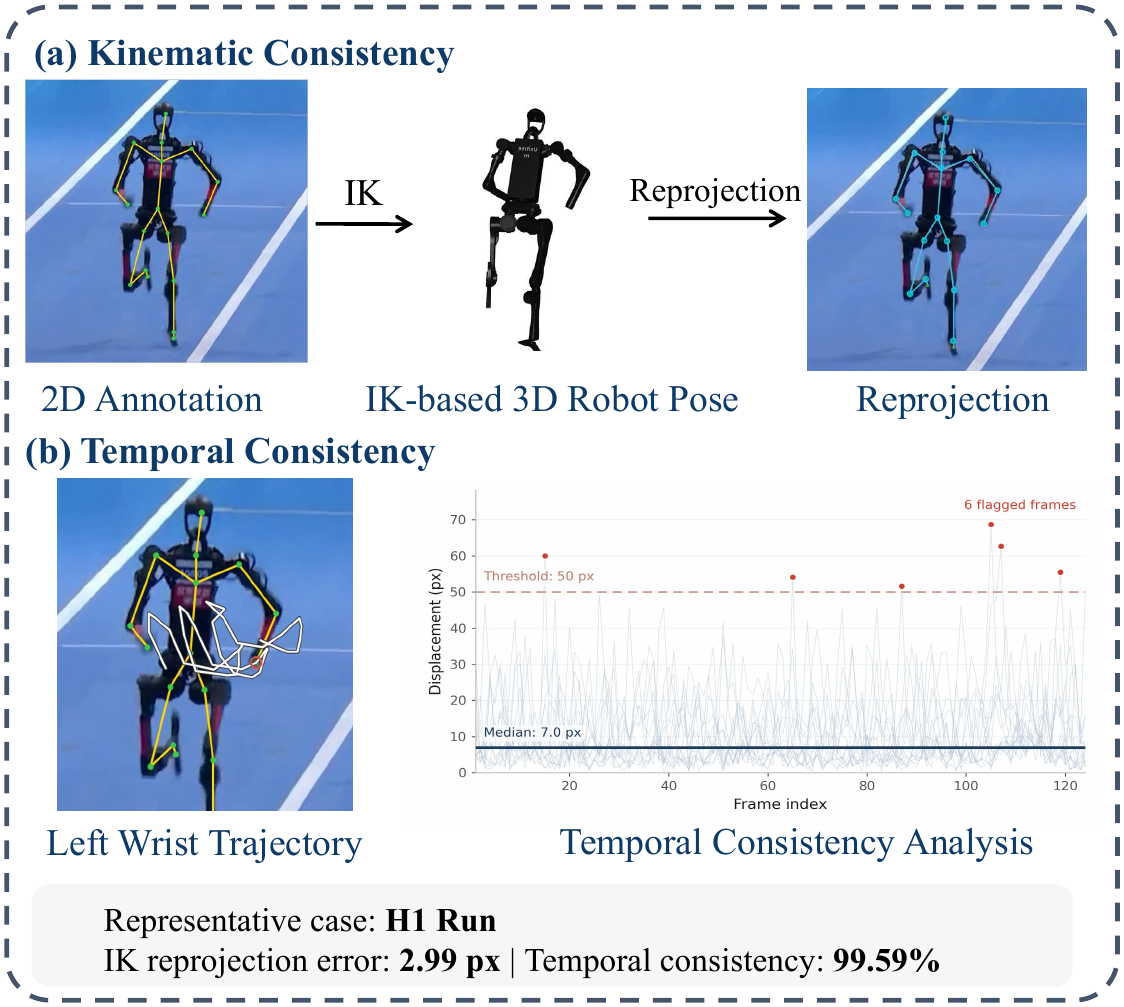}
\captionof{figure}{Real-world annotation quality verification}
\label{fig:dataset_qc}
\end{minipage}%
\hfill%
\begin{minipage}[t]{0.50\linewidth}
\vspace{0pt}
\centering

\begingroup
\captionsetup{font=small,position=top,skip=3pt}
\fontsize{7.5}{9}\selectfont
\renewcommand{\arraystretch}{1.05}
\setlength{\tabcolsep}{2pt}

\captionof{table}{Synthetic annotation quality.}
\label{tab:annotation_quality}
\begin{tabular}{@{}
    >{\raggedright\arraybackslash}m{\dimexpr0.34\linewidth-6pt\relax}
    *{3}{>{\centering\arraybackslash}m{\dimexpr0.22\linewidth-2pt\relax}}
    @{}}
\toprule
Metric & Mean & P95 & Max \\
\midrule
FK align. (mm)
& $2.83{\times}10^{-5}$ & $7.16{\times}10^{-5}$ & $2.50{\times}10^{-4}$ \\
\addlinespace[1pt]
Reproj. error (px)
& $3.40{\times}10^{-4}$ & $1.98{\times}10^{-3}$ & $1.48{\times}10^{-2}$ \\
\addlinespace[1pt]
Ground pen. (mm)
& $1.56{\times}10^{-6}$ & $1.41{\times}10^{-5}$ & $8.46{\times}10^{-5}$ \\
\addlinespace[1pt]
Joint-limit viol.\newline (deg)
& $2.01{\times}10^{-11}$ & $1.21{\times}10^{-15}$ & $4.78{\times}10^{-5}$ \\
\bottomrule
\end{tabular}

\par\vspace{5pt}

\captionof{table}{Real-world annotation quality.}

\label{tab:real_annotation_quality}

\begin{tabular}{@{}
    >{\raggedright\arraybackslash}m{\dimexpr0.34\linewidth-6pt\relax}
    *{3}{>{\centering\arraybackslash}m{\dimexpr0.22\linewidth-2pt\relax}}
    @{}}

\toprule
Metric & Mean & P95 & Temp. Consistency \\

\midrule

Keypoint velocity\newline (px/frame)
& 15.12
& 34.01
& 97.45\% \\

\addlinespace[1pt]

IK reproj.\newline (px)
& 4.29
& 15.10
& -- \\

\bottomrule

\end{tabular}

\endgroup

\end{minipage}
\end{figure}

Because annotations are obtained directly from simulator states, we verify
their consistency through FK reconstruction, camera reprojection, ground-contact
consistency, and trajectory analysis.
Table~\ref{tab:annotation_quality} summarizes the resulting statistics: the FK
alignment error is $2.83\times10^{-5}$ mm on average with a maximum of
$2.50\times10^{-4}$ mm, reprojection error stays below $1.5\times10^{-2}$ px,
ground penetration stays below $8.5\times10^{-5}$ mm, and joint-limit violations
are negligible. No hard failures were observed in the final release.

\paragraph{Real-world 2D annotations.}

For the real-world 2D annotations, no metric 3D ground truth is available, so we
evaluate their reliability through temporal and kinematic consistency. The
verified annotations exhibit low temporal
outlier rates, indicating stable keypoint trajectories across frames. We
further fit robot configurations through inverse kinematics and project the
resulting 3D poses back to the image plane. The resulting reprojection errors
show that the annotations can be largely explained by feasible robot
kinematic configurations. Table~\ref{tab:real_annotation_quality} reports the
resulting statistics: keypoint velocities average $15.12$ px/frame with a
temporal-consistency rate of $97.45\%$, and the IK reprojection error averages
$4.29$ px (P95 $15.10$ px), confirming stable keypoint trajectories and feasible
kinematic configurations.

\paragraph{Real-world 3D capture.}
For the physical Roban sequences, we report GT repeatability and calibration consistency
rather than absolute accuracy. The $23$ encoder joints exhibit $0.25^\circ$ RMS
static noise. The multi-tag root calibration shows $1$--$3$\,cm positional
inconsistency, with a median tag reprojection error of $0.24$\,px (P95
$1.63$\,px).

\subsection{Benchmark protocol}
\label{sec:benchmark_protocol}

The Humanoids benchmark supports synthetic 3D pose recovery, real-world
image-space evaluation, and real-world 3D evaluation on Roban.

For the synthetic benchmark, we provide motion-disjoint train, validation,
and test splits, such that clips derived from the same source motion are
kept within a single split. Each sample is associated with RGB observations,
camera parameters, robot states, and FK-derived 3D keypoints. The default evaluation
follows the provided robot kinematic specification and reports articulation and 3D pose metrics.

For the original real-world evaluation set, metric 3D ground truth is unavailable;
we evaluate image-space pose recovery using manually verified Canonical-18 2D
keypoints. The Roban extension instead evaluates 3D pose against recorded joint
states and estimated camera-relative root poses.

\section{HumanoidPose: Robot-Native Humanoid Pose Recovery}
\label{sec:method}

\begin{figure*}[t]
    \centering
    \includegraphics[width=\textwidth]{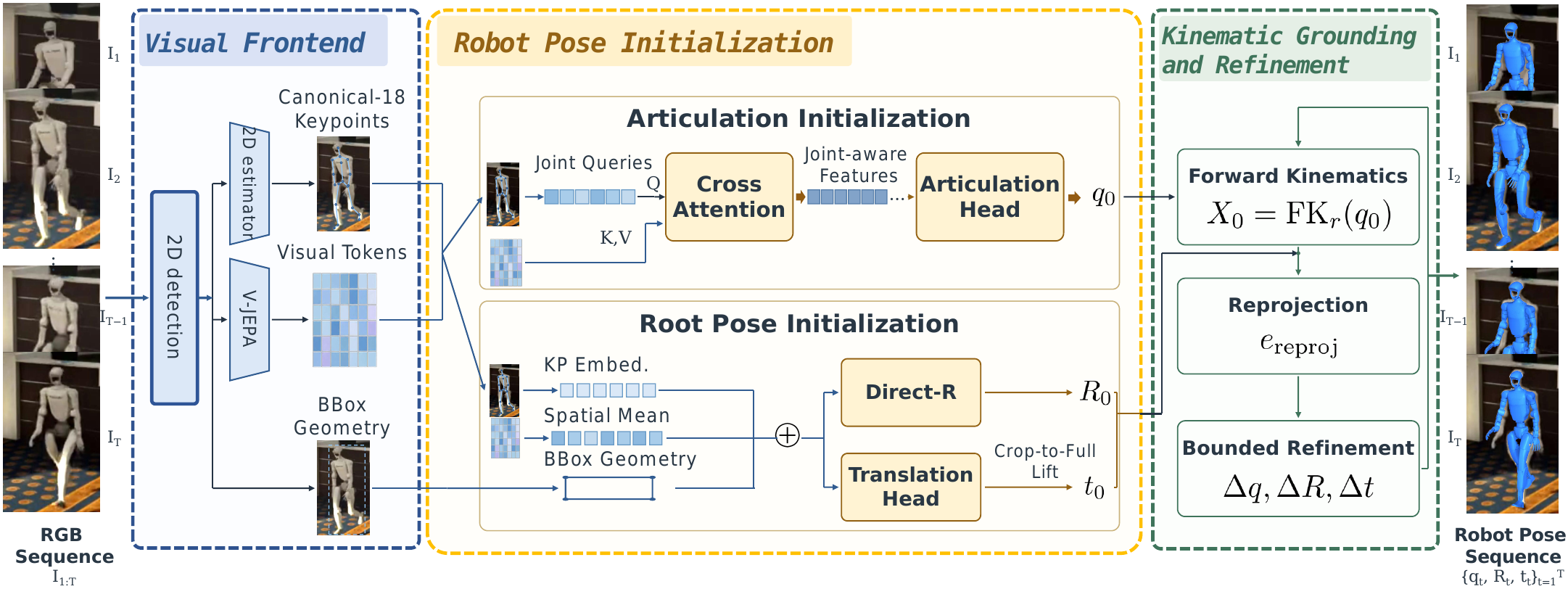}
    \caption{Overview of HumanoidPose. The visual frontend extracts Canonical-18
    keypoints, bounding-box geometry, and spatio-temporal visual tokens.
    Keypoint-conditioned joint queries initialize robot articulation $q_0$, while
    a separate root branch estimates the camera-relative root pose $(R_0,t_0)$.
    The initial state is grounded through embodiment-specific forward kinematics
    and image-space reprojection, followed by bounded residual refinement to obtain
    the final robot pose.}
    \label{fig:method_overview}
\end{figure*}

We present \textbf{HumanoidPose}, a monocular framework for recovering
robot-native articulation and camera-relative root pose. As shown in
Fig.~\ref{fig:method_overview}, it combines a frozen visual frontend
(Sec.~\ref{sec:visual_frontend}), separate articulation and root-pose
branches (Sec.~\ref{sec:pose_initialization}), and embodiment-specific
kinematic grounding with bounded refinement
(Sec.~\ref{sec:kinematic_grounding}).

\subsection{Problem Formulation}
\label{sec:problem_formulation}

Humanoid Robot Pose Estimation (HRPE) recovers robot joint configuration and
camera-relative root pose from monocular RGB.

Each humanoid embodiment $r$ is described by its kinematic model
$\mathcal{M}_r=(\operatorname{FK}_r,q_{\min},q_{\max})$, where
$\operatorname{FK}_r:\mathbb{R}^{D_r}\rightarrow\mathbb{R}^{J\times3}$
maps a joint configuration to the Canonical-18 keypoints
(Appendix~\ref{app:canonical18}) in the robot root frame. Here, $D_r$ is the
number of robot joints and $J$ is the number of keypoints. Given a monocular
RGB sequence window $I_{1:T}$, HRPE estimates
\begin{equation}
    (q,R,t)
    =
    \mathcal{F}_{\theta_r}(I_{1:T};\mathcal{M}_r),
    \label{eq:hrpe}
\end{equation}
where $q\in[q_{\min},q_{\max}]$ denotes the joint configuration at the target
time step and $(R,t)\in\mathrm{SO}(3)\times\mathbb{R}^3$ denotes the
camera-relative root pose. The corresponding 3D robot structure is obtained
through $\operatorname{FK}_r(q)$. We use the same architecture for all
embodiments, but train separate model parameters for each embodiment.

HRPE requires handling image-space displacement and scale variation from
floating-base motion, extracting local cues for high-DoF articulation under
self-occlusion, and respecting embodiment-specific joint definitions and
kinematics. HumanoidPose addresses these challenges through human-inspired
visual perception, robot-native pose estimation, and kinematic grounding.

\subsection{Human-Inspired Visual Perception}
\label{sec:visual_frontend}

Given an RGB sequence, a humanoid detector~\citep{redmon2016you} provides a
target bounding box $b$. The resulting target-centric crops reduce sensitivity
to image-space position and scale variation, while the bounding-box geometry
supports camera-relative root estimation.

A 2D pose estimator~\citep{khirodkarsapiens2}, pre-adapted on synthetic
humanoid data, predicts Canonical-18 keypoints
$\hat{x}\in\mathbb{R}^{J\times2}$ with confidence scores
$c\in\mathbb{R}^{J}$. These human-inspired keypoints provide semantic body-part
cues, while the final joint configuration remains robot-native. A frozen
V-JEPA encoder~\citep{bardes2024revisiting} extracts spatio-temporal visual
tokens $z$ from the target crops. All frontend modules remain fixed during HRPE
training. Together, $z$, $\hat{x}$, $c$, and $b$ provide complementary
visual, structural, and geometric cues.

\subsection{Robot-Native Pose Estimation}
\label{sec:pose_initialization}

Articulation estimation relies primarily on local body configurations and
fine-grained visual cues, whereas camera-relative root pose depends more
strongly on global orientation, image position, scale, and geometry. We
therefore initialize them through separate branches:
\begin{equation}
    q_0 = f_q(z,\hat{x};r),\qquad
    R_0 = f_R(z,\hat{x},b;r),\qquad
    t_0 = f_t(z,\hat{x},b;r).
    \label{eq:pose_init}
\end{equation}
The keypoint inputs also carry their confidence scores.

\paragraph{Articulation branch.}
Directly regressing high-DoF joint states from a global representation may
underuse local body-part evidence. We instead construct queries conditioned on
the Canonical-18 keypoints and combine them with visual tokens through
cross-attention~\citep{vaswani2017attention}:
$h=\operatorname{CrossAttn}\bigl(Q(\hat{x}),z\bigr)$.
Each query is associated with a semantic keypoint and retrieves
articulation-relevant visual evidence. An articulation decoder maps the
resulting features to the embodiment-specific joint configuration $q_0$.
Thus, keypoints provide a structured interface for robot-native prediction.

\paragraph{Root branch.}
The root branch combines visual features, Canonical-18 keypoints, and
bounding-box geometry to initialize camera-relative pose. It directly predicts
root rotation and estimates root translation, using the crop-to-full-image
coordinate transformation to account for target location and scale. The
resulting state $(q_0,R_0,t_0)$ is subsequently grounded in the robot
kinematic model.

\subsection{Kinematic Grounding and Pose Refinement}
\label{sec:kinematic_grounding}

Although human-inspired keypoints provide structural visual cues, the
recovered pose must follow the robot's own joint definitions and kinematics.
We therefore apply embodiment-specific forward kinematics to the initialized
articulation, obtaining
$X_0=\operatorname{FK}_{r}(q_0)$,
the metric 3D robot structure in the root frame.

The FK-derived structure is projected into camera and image space, and its
reprojection residuals with respect to the observed keypoints provide
geometric cues for refinement. The refinement module predicts bounded residual
corrections to $(q_0,R_0,t_0)$, yielding the final pose $(q,R,t)$ while
restricting changes around the initial estimate. Joint-limit constraints
maintain embodiment-specific feasibility. Feature construction and residual
parameterization are detailed in Appendix~\ref{app:kinematic_refinement}.

\subsection{Training Objectives}
\label{sec:training_objective}

Given the ground-truth state $(q^*,R^*,t^*)$, HumanoidPose is supervised in
joint, Cartesian, camera, and image spaces:
\begin{equation}
    \mathcal{L}
    =
    \lambda_q\mathcal{L}_q
    +\lambda_{\mathrm{FK}}\mathcal{L}_{\mathrm{FK}}
    +\lambda_R\mathcal{L}_R
    +\lambda_t\mathcal{L}_t
    +\lambda_{\mathrm{cam}}\mathcal{L}_{\mathrm{cam}}
    +\lambda_{\mathrm{repr}}\mathcal{L}_{\mathrm{repr}}
    +\lambda_{\Delta}\mathcal{L}_{\Delta}.
    \label{eq:objective}
\end{equation}
For robot-state supervision, $\mathcal{L}_q$ and $\mathcal{L}_t$ are
Smooth-$\ell_1$ losses, and $\mathcal{L}_R$ is a geodesic rotation loss.
The FK loss
$\mathcal{L}_{\mathrm{FK}}
=
\|\operatorname{FK}_r(q)-\operatorname{FK}_r(q^*)\|_1$
supervises 3D structure, while $\mathcal{L}_{\mathrm{cam}}$ enforces
camera-space consistency. The reprojection loss
$\mathcal{L}_{\mathrm{repr}}$ enforces image-space consistency, and
$\mathcal{L}_{\Delta}$ regularizes residual corrections. Detailed loss
definitions are provided in Appendix~\ref{app:refinement_loss}.

\section{Experiments}
\label{sec:experiments}

\subsection{Experimental Setup}
\label{sec:exp_setup}
Following Sec.~\ref{sec:benchmark_protocol}, we train and evaluate on the
synthetic split of \textbf{Humanoids} and assess sim-to-real transfer on
\textbf{Humanoids-Real} and external \textbf{DHRP-H1}~\citep{heo2024diverse}.
We additionally evaluate real-world 3D pose on a physical Roban humanoid.
All real-world evaluations are \emph{zero-shot}, with no real-image adaptation
or fine-tuning. Each embodiment uses the same architecture with independently
trained parameters; the benchmark measures per-embodiment accuracy.
The main synthetic and real-world 2D comparisons report macro averages over
the original 12 embodiments; Roban 3D results are reported separately.
Per-embodiment results are provided in the appendix.
For synthetic evaluation and Roban real-world 3D evaluation, we report
\textbf{MPJAE}, \textbf{MPJPE}, \textbf{PA-MPJPE}, \textbf{G-MPJPE},
and \textbf{PCK@0.3}. For the real-world 2D datasets, we report
image-space metrics, including \textbf{2D MPJPE},
\textbf{NME}, and \textbf{PCK@0.1}. Detailed metric definitions are provided in
Appendix~\ref{sec:metrics}.

\subsection{Comparison with Existing Methods}
\label{sec:comparison}

We compare representative HPE and RPE methods under the HRPE protocol.
HPE baselines use either (1) direct application with human-to-robot retargeting
(e.g., GMR~\citep{araujo2025retargeting}) without additional training, or
(2) robot-specific output representations trained on Humanoids.
RPE baselines retain their original formulations, with outputs adapted to each
humanoid embodiment and training on its corresponding synthetic data.
Implementation details are in Appendix~\ref{sec:baseline_details}.

\begin{table}[t]
\centering
\scriptsize

\caption{
\textbf{Comparison with existing pose estimation methods on Humanoids-3D (macro average over the original 12 embodiments).}
$^{\star}$ denotes humanoid-specific adaptation while preserving the original
inference pipeline.
}
\label{tab:sim_pose_estimation}

\vspace{1mm}

\setlength{\tabcolsep}{2pt}
\renewcommand{\arraystretch}{0.88}

\resizebox{0.94\linewidth}{!}{
\begin{tabular}{c|l|ccccc}
\toprule
Category
& Method
& MPJAE$\downarrow$
& MPJPE$\downarrow$
& PA-MPJPE$\downarrow$
& G-MPJPE$\downarrow$
& PCK@0.3$\uparrow$
\\
\midrule

\multirow{6}{*}{HPE}
& HMR2~\citep{goel2023humans} + GMR
& 29.503 & 150.772 & 115.508 & 2909.486 & 65.04
\\
& GVHMR~\citep{shen2024world} + GMR
& 28.471 & 140.666 & 113.203 & 2598.000 & 70.78
\\
& WHAM~\citep{shin2024wham} + GMR
& 32.754 & 189.039 & 163.659 & 2980.960 & 62.24
\\
& HMR2$^{\star}$~\citep{goel2023humans}
& 13.634 & 75.574 & 67.632 & 320.149 & 84.74
\\
& GVHMR$^{\star}$~\citep{shen2024world}
& 13.522 & 75.606 & 68.011 & 331.569 & 84.81
\\
& WHAM$^{\star}$~\citep{shin2024wham}
& 16.902 & 100.052 & 91.797 & 803.063 & 77.88
\\

\midrule

\multirow{3}{*}{RPE}
& RoboPEPP$^{\star}$~\citep{goswami2025robopepp}
& 15.771 & 89.200 & 80.230 & 2501.417 & 80.81
\\
& HoRoPose$^{\star}$~\citep{ban2024real}
& 22.949 & 145.910 & 135.039 & 1346.556 & 69.72
\\
& RoboPose$^{\star}$~\citep{labbe2021single}
& 16.139 & 91.173 & 81.125 & 540.262 & 80.45
\\

\midrule

HRPE
& \textbf{HumanoidPose (Ours)}
& \textbf{12.430}
& \textbf{59.954}
& \textbf{52.171}
& \textbf{183.218}
& \textbf{89.75}
\\

\bottomrule
\end{tabular}
}

\vspace{2mm}
\end{table}

\begin{figure*}[t]
    \centering

    \includegraphics[width=\textwidth]{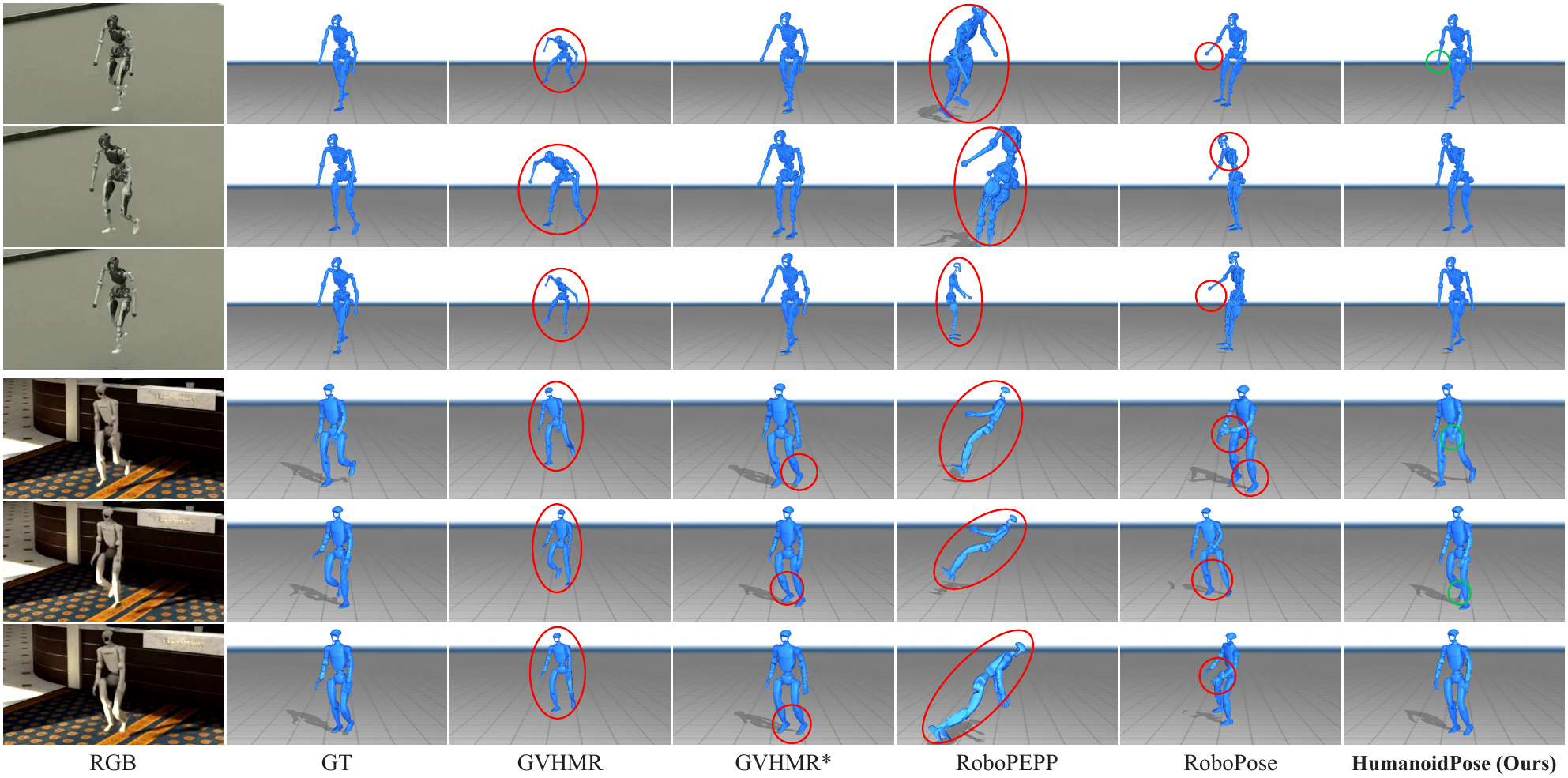}
    \caption{
    Visual comparison of HumanoidPose against baseline methods on the synthetic Humanoids benchmark.
Results are shown for G1 (bottom) and PNDAdamLite (top).
Here $\star$ denotes humanoid-specific adaptation.}
    \label{fig:sim3d_qualitative}
    \vspace{-12pt}
\end{figure*}
\vspace{-2mm}

As shown in Table~\ref{tab:sim_pose_estimation}, HumanoidPose consistently
achieves strong performance across articulation estimation, root-relative 3D
pose recovery, and global camera-space evaluation. The comparison highlights
two observations. First, direct transfer of HPE models through human-to-robot
retargeting suffers from substantial performance degradation, whereas
robot-specific adaptation improves their accuracy, indicating that
human-centric visual priors require alignment with robot-specific structures.
Second, existing RPE methods provide robot-aware representations but are
mainly designed for constrained robot systems. HumanoidPose addresses both
limitations by directly modeling robot articulation with explicit kinematic
grounding.

Figure~\ref{fig:sim3d_qualitative} further illustrates these differences.
GVHMR suffers from human-body prior mismatch, particularly for smaller
humanoid platforms such as Unitree G1, while RoboPEPP's geometry-based root
estimation can degrade on humanoid robots with diverse embodiments.
HumanoidPose better preserves robot articulation and camera-relative root pose.

\subsection{Real-World Evaluation}
\label{sec:real-world}

Table~\ref{tab:real2d_pose_estimation} compares HPE and RPE baselines on
\textbf{Humanoids-Real} and \textbf{DHRP-H1}. As metric 3D robot states are
unavailable, we evaluate projected predictions against manual 2D annotations.

HumanoidPose improves image-space pose accuracy on both benchmarks.
The results also indicate meaningful transfer from HPE models, with
robot-specific adaptation and explicit kinematics further improving joint
localization.

For real-world 3D evaluation, we record 24 Roban sequences (23,170 frames)
with a calibrated RGB camera and obtain joint states from the robot SDK and
camera-relative root poses from torso-mounted AprilTags. HumanoidPose and the
adapted baselines are trained on the training split of 574 synthetic Roban
clips (60,232 frames) and are evaluated on real videos without real-world fine-tuning. As shown in
Table~\ref{tab:real3d_roban}, HumanoidPose achieves 13.65$^\circ$ MPJAE,
42.82 mm MPJPE, and 92.13\% PCK@0.3, with the best result across all five
metrics. Further collection details are in Appendix~\ref{app:roban}. Synthetic Roban results are reported in Appendix~\ref{app:roban_synthetic}.

\subsection{Ablation Studies}
\label{sec:ablation}

{
\begin{table}[t]

\centering

\begin{minipage}[t]{0.48\linewidth}
\centering

\captionof{table}{Real-world 3D evaluation on Roban. $^{\star}$ denotes humanoid-specific adaptation.}
\label{tab:real3d_roban}
\vspace{2mm}

\scriptsize

\setlength{\tabcolsep}{1.5pt}

\renewcommand{\arraystretch}{1.12}

\resizebox{\linewidth}{!}{
\begin{tabular}{lccccc}
\toprule
Method
& MPJAE$\downarrow$
& MPJPE$\downarrow$
& PA-MPJPE$\downarrow$
& G-MPJPE$\downarrow$
& PCK@0.3$\uparrow$
\\
\midrule

\multicolumn{6}{@{}l}{\itshape HPE} \\
HMR2+GMR & 33.97 & 84.89 & 82.11 & 403.83 & 76.90 \\
GVHMR+GMR & 24.30 & 55.17 & 51.76 & 245.00 & 89.89 \\
WHAM+GMR & 44.89 & 141.07 & 139.20 & 656.98 & 71.75 \\
HMR2$^{\star}$ & 18.22 & 56.56 & 55.90 & 535.34 & 85.85 \\
GVHMR$^{\star}$ & 20.77 & 75.12 & 73.62 & 624.80 & 83.65 \\
WHAM$^{\star}$ & 17.06 & 62.50 & 61.25 & 733.23 & 84.70 \\
\midrule

\multicolumn{6}{@{}l}{\itshape RPE} \\
RoboPEPP$^{\star}$ & 14.87 & 55.91 & 55.09 & 3004.89 & 86.47 \\
HoRoPose$^{\star}$ & 19.93 & 103.03 & 99.32 & 854.85 & 76.40 \\
RoboPose$^{\star}$ & 16.33 & 52.74 & 52.04 & 625.46 & 89.28 \\
\midrule

\multicolumn{6}{@{}l}{\itshape HRPE} \\
\textbf{HumanoidPose} & \textbf{13.65} & \textbf{42.82} & \textbf{42.17}
& \textbf{199.83} & \textbf{92.13} \\

\bottomrule
\end{tabular}}

\end{minipage}\hfill
\begin{minipage}[t]{0.50\linewidth}
\centering

\captionof{table}{Real-world 2D evaluation. $^{\star}$ denotes humanoid-specific adaptation.}
\label{tab:real2d_pose_estimation}
\vspace{2mm}

\scriptsize

\setlength{\tabcolsep}{1.5pt}

\renewcommand{\arraystretch}{1.0}

\resizebox{\linewidth}{!}{
\begin{tabular}{lcccccc}
\toprule
\multirow{2}{*}{Method}
& \multicolumn{3}{c}{Humanoids-Real}
& \multicolumn{3}{c}{DHRP-H1} \\
\cmidrule(lr){2-4}\cmidrule(lr){5-7}
& MPJPE$\downarrow$ & NME$\downarrow$ & PCK@0.1$\uparrow$
& MPJPE$\downarrow$ & NME$\downarrow$ & PCK@0.1$\uparrow$ \\
\midrule
\multicolumn{7}{@{}l}{\itshape HPE} \\
GVHMR+GMR & 78.82 & 0.1371 & 53.27 & 97.62 & 0.1470 & 62.06 \\
HMR2+GMR & 247.77 & 0.4067 & 5.77 & 253.68 & 0.3285 & 10.61 \\
WHAM+GMR & 197.97 & 0.3098 & 15.81 & 491.47 & 0.6015 & 0.99 \\
GVHMR$^{\star}$ & 149.27 & 0.2432 & 22.97 & 98.81 & 0.1481 & 60.69 \\
HMR2$^{\star}$ & 162.98 & 0.2607 & 15.94 & 146.48 & 0.1941 & 15.73 \\
WHAM$^{\star}$ & 243.53 & 0.4048 & 7.49 & 406.34 & 0.5003 & 2.06 \\
\midrule
\multicolumn{7}{@{}l}{\itshape RPE} \\
RoboPEPP$^{\star}$ & 139.18 & 0.2157 & 23.44 & 282.25 & 0.3684 & 8.40 \\
HoRoPose$^{\star}$ & 128.32 & 0.2023 & 28.72 & 208.18 & 0.2830 & 17.48 \\
RoboPose$^{\star}$ & 152.10 & 0.2356 & 20.64 & 212.69 & 0.2901 & 14.73 \\
\midrule
\multicolumn{7}{@{}l}{\itshape HRPE} \\
\textbf{HumanoidPose} & \textbf{71.49} & \textbf{0.1180} & \textbf{61.97}
& \textbf{87.80} & \textbf{0.1363} & \textbf{74.20} \\
\bottomrule
\end{tabular}}

\end{minipage}

\end{table}
}
{
\begin{table}[t]

\centering

\begin{minipage}[t]{0.75\linewidth}
\centering

\captionof{table}{Ablation study of HumanoidPose components on Humanoids-3D.}
\label{tab:main_ablation}
\vspace{2mm}

\scriptsize

\setlength{\tabcolsep}{1.5pt}

\renewcommand{\arraystretch}{1.0}

\resizebox{\linewidth}{!}{
\begin{tabular}{lccccc}
\toprule
Variant
& MPJAE$\downarrow$
& MPJPE$\downarrow$
& PA-MPJPE$\downarrow$
& G-MPJPE$\downarrow$
& PCK@0.3$\uparrow$ \\
\midrule

\textit{w/o} Robot Artic.
& 14.668 & 72.237 & 61.115 & 190.705 & 86.87 \\

\shortstack[l]{\textit{w/o} Decoup. Root}
& 12.537 & 60.429 & 52.664 & 213.423 & 89.51 \\

\shortstack[l]{\textit{w/o} Kinematic\\
Grounding \& Refinement}
& 12.851 & 64.705 & 56.064 & 194.082 & 88.26 \\

\midrule

\textbf{Full HumanoidPose}
& \textbf{12.430}
& \textbf{59.954}
& \textbf{52.171}
& \textbf{183.218}
& \textbf{89.75} \\

\bottomrule
\end{tabular}}

\end{minipage}

\end{table}
}

We conduct ablation studies on Humanoids-3D to analyze the contribution of
different components in HumanoidPose. As shown in Table~\ref{tab:main_ablation},
removing robot-native articulation causes the largest degradation, highlighting
the importance of predicting robot-specific joint states. Kinematic grounding and refinement further improve performance by incorporating
robot-specific geometric constraints. The decoupled
articulation and root pose initialization, together with kinematic grounding,
mainly improves global camera-space alignment, as indicated by the increase in
G-MPJPE after removal. A more comprehensive set of ablations, including
visual-frontend and standard-regressor comparisons, is provided in
Appendix~\ref{app:additional_ablation}.

We further evaluate the robustness to different perception frontends, including
the V-JEPA~\citep{bardes2024revisiting} and DINOv2~\citep{oquab2023dinov2}
encoders and the Sapiens2~\citep{khirodkarsapiens2} and
ViTPose-H~\citep{xu2022vitpose} 2D pose frontends, in
Table~\ref{tab:frontend_robustness}. Replacing either the visual encoder or the
2D pose frontend maintains comparable performance, indicating that HumanoidPose
does not rely on a specific frontend configuration.

\section{Conclusion}

We introduced \emph{Humanoid Robot Pose Estimation (HRPE)}, a monocular perception
task that aims to recover both robot-native articulation and camera-relative
free-base pose for humanoid robots. We further constructed a benchmark with
synthetic and real-world data across 13 humanoid embodiments, including
Roban synthetic data and real-world 3D evaluation. We also proposed HumanoidPose, combining visual perception,
robot-native articulation prediction, asymmetric root initialization, and
kinematic grounding.

Experiments demonstrate the effectiveness of robot-native articulation modeling and kinematic grounding for monocular 3D humanoid pose estimation.

\section*{\textsc{AI Use Statement}}

\small
We acknowledge the use of generative AI tools to assist in improving the
manuscript's language clarity, refining the presentation of experimental
procedures, providing preliminary insights for empirical observations, and
offering structural suggestions for organizing figures, tables, and overall
manuscript layout.

The authors maintained full scientific oversight throughout the entire
process. All AI-generated suggestions were carefully reviewed and revised
where necessary. All experimental implementations, empirical evaluations,
and literature references were independently verified by the research team.

The authors take full responsibility for the accuracy, integrity, and
conclusions of the work presented in this manuscript.
\normalsize

\section*{\textsc{Ethics Statement}}

\small
This work studies monocular perception of humanoid robots and does not
involve the collection or processing of personally identifiable information.
The proposed synthetic benchmark is generated in simulation using robot
models and motion data, while real-world evaluations focus on humanoid robot
observations without human subjects or identity-related annotations.

We acknowledge that visual perception capabilities for humanoid robots may
serve as a foundation for future human--robot interaction and autonomous
robotic systems. Such technologies should be developed and deployed with
careful consideration of safety, reliability, and appropriate human oversight.
This work aims to facilitate research on robot state understanding through
open evaluation protocols and benchmarks.
\normalsize

{\small
\bibliographystyle{iclr2027_conference}
\bibliography{iclr2027_conference}
}

\clearpage
\appendix

\section*{Appendices}

\section{Frontend Robustness Analysis}
\label{app:frontend_robustness}

\begin{table}[ht]
\centering
\begin{minipage}{0.75\linewidth}
\centering
\captionof{table}{Frontend robustness analysis. $\downarrow$ lower is better; $\uparrow$ higher is better.}
\label{tab:frontend_robustness}
\vspace{2mm}

\scriptsize

\setlength{\tabcolsep}{1.5pt}

\renewcommand{\arraystretch}{1.0}

\resizebox{\linewidth}{!}{
\begin{tabular}{lccccc}
\toprule
Frontend & MPJAE$\downarrow$ & MPJPE$\downarrow$ & PA-MPJPE$\downarrow$ & G-MPJPE$\downarrow$ & PCK@0.3$\uparrow$ \\
\midrule
V-JEPA+Sapiens2 & 12.430 & 59.954 & 52.171 & 183.218 & 89.75 \\
DINOv2+Sapiens2 & 11.497 & 55.972 & 49.199 & 199.485 & 90.59 \\
V-JEPA+ViTPose-H & 11.998 & 57.037 & 49.448 & 180.049 & 90.42 \\
DINOv2+ViTPose-H & 11.444 & 55.245 & 48.468 & 197.274 & 91.05 \\
\bottomrule
\end{tabular}}

\end{minipage}
\end{table}

\section{Additional Ablation Studies}
\label{app:additional_ablation}

Table~\ref{tab:additional_ablation} reports a more comprehensive ablation study
that complements the main ablation table. The six configurations answer four
questions. (i) \emph{Visual frontend} (two runs): removing keypoint conditioning
or the robot-detection guidance probes the contribution of the 2D keypoint
conditioning and of target localization. (ii) \emph{Kinematic module} (two
runs): removing kinematic grounding or the bounded refinement isolates the
contribution of the kinematic constraint and of the refinement module. (iii)
\emph{Architecture comparison} (two runs): a standard regressor with FK, with
and without the predicted 2D keypoints, provides a baseline that does not use
joint queries. (iv) \emph{Joint-query contribution}: comparing the standard regressor
that uses the same predicted 2D keypoints (63.761) with the joint-query model
without the bounded refinement (60.260) shows that the joint-query decoder
provides an additional gain beyond the 2D keypoints themselves. Note that
\textit{w/o} Bounded Refinement coincides with \textit{w/o} Refinement in the
main ablation table and is counted only once.

\begin{table}[t]
\centering
\small

\caption{
\textbf{Additional ablation studies.} All metrics are reported on Humanoids-3D.
$\downarrow$ lower is better; $\uparrow$ higher is better.
}
\label{tab:additional_ablation}

\vspace{1mm}

\setlength{\tabcolsep}{3pt}
\renewcommand{\arraystretch}{1.0}

\resizebox{0.94\linewidth}{!}{
\begin{tabular}{lccccc}
\toprule
Variant & MPJAE$\downarrow$ & MPJPE$\downarrow$ & PA-MPJPE$\downarrow$
& G-MPJPE$\downarrow$ & PCK\@0.3$\uparrow$ \\
\midrule

\multicolumn{6}{@{}l}{\itshape Visual Frontend} \\
\textit{w/o} Keypoint Conditioning
& 16.974 & 90.737 & 80.694 & 207.315 & 80.69 \\
\textit{w/o} Humanoid Detection Guided
& 17.826 & 97.139 & 86.698 & 475.152 & 78.70 \\
\midrule

\multicolumn{6}{@{}l}{\itshape Kinematic Modeling} \\
\textit{w/o} Kinematic Grounding
& 12.770 & 64.015 & 55.625 & 185.713 & 88.37 \\
\textit{w/o} Bounded Refinement
& 12.503 & 60.260 & 52.353 & 191.579 & 89.62 \\
\midrule

\multicolumn{6}{@{}l}{\itshape Architecture Comparison} \\
Standard Regressor + FK
& 16.099 & 86.422 & 77.898 & 385.970 & 81.59 \\
Standard Regressor + FK + Predicted 2D KP
& 13.138 & 63.761 & 53.376 & 203.788 & 88.39 \\
\midrule

\textbf{Full HumanoidPose}
& \textbf{12.430} & \textbf{59.954} & \textbf{52.171} & \textbf{183.218} & \textbf{89.75} \\

\bottomrule
\end{tabular}}
\end{table}

\section{Per-Robot Breakdown of the Main Experiments}
\label{sec:perrobot}

Tables~\ref{tab:perrobot_sim_mpjae_deg}--\ref{tab:perrobot_sim_pck_0p3}
break down every Humanoids-3D method of the main comparison by robot, and
Tables~\ref{tab:perrobot_real2d_2d_mpjpe_px}--\ref{tab:perrobot_real2d_pck0p10}
do the same on the internal Humanoids-Real 2D set.  In each table the
\emph{Mean} column is the robot-balanced average, i.e.\ the value reported in
the main comparison tables; the per-robot entries are computed under the shared
evaluation protocol of Section~\ref{sec:baseline_details}.  The external DHRP-H1
benchmark is a single image set and is therefore not broken down by robot.

\begin{table}[!htbp]
\centering
\scriptsize
\setlength{\tabcolsep}{3pt}
\renewcommand{\arraystretch}{0.9}
\caption{Per-robot MPJAE ($^\circ$) on Humanoids-3D. Robots: H1=UnitreeH1, G1=G1, BT1=BoosterT1, BK1=BoosterK1, BHL=BerkeleyHumanoidLite, EPM=EngineAI\_PM01, FGR3=FourierGR3, FN1=FourierN1, KS45=KuavoS45, PND=PNDAdamLite, STD=StanfordToddy, TK=TienKung. The \emph{Mean} column is the robot-balanced average reported in the main table.}
\label{tab:perrobot_sim_mpjae_deg}
\vspace{1mm}
\resizebox{\linewidth}{!}{
\begin{tabular}{l|cccccccccccc|c}
\toprule
Method & H1 & G1 & BT1 & BK1 & BHL & EPM & FGR3 & FN1 & KS45 & PND & STD & TK & Mean \\
\midrule
HMR2 + GMR & 32.1 & 38.8 & 12.9 & 41.9 & 31.7 & 48.8 & 27.7 & 30.5 & 24.1 & 22.3 & 16.1 & 27.0 & \textbf{29.5} \\
GVHMR + GMR & 48.8 & 24.0 & 10.2 & 39.4 & 21.0 & 41.2 & 33.4 & 31.1 & 34.9 & 21.8 & 17.1 & 18.7 & \textbf{28.5} \\
WHAM + GMR & 44.4 & 42.6 & 16.7 & 52.2 & 31.0 & 48.7 & 23.0 & 21.1 & 28.3 & 24.7 & 16.5 & 43.6 & \textbf{32.8} \\
HMR2$^{\star}$ & 16.6 & 13.4 & 9.7 & 14.0 & 14.7 & 17.0 & 11.2 & 13.2 & 14.9 & 12.3 & 11.8 & 14.8 & \textbf{13.6} \\
GVHMR$^{\star}$ & 16.4 & 13.6 & 8.9 & 14.1 & 14.8 & 16.8 & 11.4 & 13.5 & 14.6 & 12.1 & 12.1 & 14.0 & \textbf{13.5} \\
WHAM$^{\star}$ & 16.0 & 17.7 & 13.6 & 17.1 & 15.5 & 23.3 & 15.4 & 17.2 & 20.3 & 13.4 & 13.3 & 20.0 & \textbf{16.9} \\
RoboPEPP$^{\star}$ & 16.4 & 16.2 & 11.1 & 16.3 & 16.1 & 22.0 & 13.2 & 16.0 & 18.4 & 12.9 & 12.8 & 17.9 & \textbf{15.8} \\
HoRoPose$^{\star}$ & 19.3 & 32.6 & 18.3 & 20.5 & 19.8 & 26.6 & 18.4 & 18.8 & 22.4 & 15.6 & 26.3 & 36.8 & \textbf{22.9} \\
RoboPose$^{\star}$ & 16.3 & 18.8 & 12.0 & 17.5 & 15.6 & 22.1 & 13.8 & 16.7 & 17.4 & 13.7 & 12.5 & 17.3 & \textbf{16.1} \\
\textbf{HumanoidPose (Ours)} & 14.2 & 13.0 & 7.7 & 13.8 & 14.5 & 16.3 & 11.1 & 13.5 & 11.2 & 10.4 & 10.6 & 12.8 & \textbf{12.4} \\
\bottomrule
\end{tabular}}
\vspace{2mm}
\end{table}

\begin{table}[!htbp]
\centering
\scriptsize
\setlength{\tabcolsep}{3pt}
\renewcommand{\arraystretch}{0.9}
\caption{Per-robot MPJPE (mm) on Humanoids-3D. Robots: H1=UnitreeH1, G1=G1, BT1=BoosterT1, BK1=BoosterK1, BHL=BerkeleyHumanoidLite, EPM=EngineAI\_PM01, FGR3=FourierGR3, FN1=FourierN1, KS45=KuavoS45, PND=PNDAdamLite, STD=StanfordToddy, TK=TienKung. The \emph{Mean} column is the robot-balanced average reported in the main table.}
\label{tab:perrobot_sim_mpjpe_mm}
\vspace{1mm}
\resizebox{\linewidth}{!}{
\begin{tabular}{l|cccccccccccc|c}
\toprule
Method & H1 & G1 & BT1 & BK1 & BHL & EPM & FGR3 & FN1 & KS45 & PND & STD & TK & Mean \\
\midrule
HMR2 + GMR & 137.4 & 187.2 & 47.8 & 80.6 & 133.9 & 273.6 & 245.6 & 146.7 & 139.5 & 210.1 & 45.6 & 161.3 & \textbf{150.8} \\
GVHMR + GMR & 207.0 & 144.1 & 46.4 & 86.9 & 68.7 & 160.4 & 297.2 & 141.7 & 233.7 & 159.1 & 48.1 & 94.8 & \textbf{140.7} \\
WHAM + GMR & 255.2 & 280.4 & 59.5 & 104.4 & 142.6 & 284.2 & 169.5 & 117.3 & 272.1 & 210.5 & 42.5 & 330.2 & \textbf{189.0} \\
HMR2$^{\star}$ & 77.9 & 83.5 & 38.2 & 38.4 & 54.6 & 128.3 & 103.6 & 71.9 & 103.7 & 94.0 & 34.4 & 78.4 & \textbf{75.6} \\
GVHMR$^{\star}$ & 77.9 & 88.6 & 34.3 & 38.4 & 55.4 & 129.8 & 107.6 & 72.2 & 103.6 & 91.0 & 35.5 & 73.0 & \textbf{75.6} \\
WHAM$^{\star}$ & 75.2 & 121.1 & 50.7 & 47.2 & 61.6 & 158.6 & 159.5 & 97.0 & 161.3 & 106.1 & 42.2 & 120.1 & \textbf{100.1} \\
RoboPEPP$^{\star}$ & 77.1 & 111.9 & 40.2 & 42.4 & 59.8 & 145.2 & 128.6 & 88.1 & 142.4 & 98.9 & 39.2 & 96.5 & \textbf{89.2} \\
HoRoPose$^{\star}$ & 111.1 & 226.3 & 75.8 & 64.5 & 69.2 & 189.4 & 192.9 & 101.0 & 174.5 & 131.2 & 57.6 & 357.5 & \textbf{145.9} \\
RoboPose$^{\star}$ & 74.7 & 133.4 & 43.4 & 50.0 & 57.4 & 147.6 & 138.0 & 94.7 & 118.8 & 101.9 & 38.8 & 95.3 & \textbf{91.2} \\
\textbf{HumanoidPose (Ours)} & 55.9 & 73.6 & 24.9 & 32.5 & 50.8 & 100.6 & 91.4 & 66.4 & 69.3 & 70.3 & 23.6 & 60.2 & \textbf{60.0} \\
\bottomrule
\end{tabular}}
\vspace{2mm}
\end{table}

\begin{table}[!htbp]
\centering
\scriptsize
\setlength{\tabcolsep}{3pt}
\renewcommand{\arraystretch}{0.9}
\caption{Per-robot PA-MPJPE (mm) on Humanoids-3D. Robots: H1=UnitreeH1, G1=G1, BT1=BoosterT1, BK1=BoosterK1, BHL=BerkeleyHumanoidLite, EPM=EngineAI\_PM01, FGR3=FourierGR3, FN1=FourierN1, KS45=KuavoS45, PND=PNDAdamLite, STD=StanfordToddy, TK=TienKung. The \emph{Mean} column is the robot-balanced average reported in the main table.}
\label{tab:perrobot_sim_pa_mpjpe_mm}
\vspace{1mm}
\resizebox{\linewidth}{!}{
\begin{tabular}{l|cccccccccccc|c}
\toprule
Method & H1 & G1 & BT1 & BK1 & BHL & EPM & FGR3 & FN1 & KS45 & PND & STD & TK & Mean \\
\midrule
HMR2 + GMR & 128.7 & 120.6 & 47.1 & 80.6 & 134.6 & 151.5 & 168.7 & 103.6 & 145.4 & 129.5 & 37.0 & 139.0 & \textbf{115.5} \\
GVHMR + GMR & 209.4 & 102.5 & 37.6 & 80.7 & 63.2 & 96.7 & 207.8 & 87.4 & 228.4 & 130.2 & 37.7 & 76.9 & \textbf{113.2} \\
WHAM + GMR & 244.1 & 193.5 & 61.8 & 96.2 & 145.8 & 208.0 & 147.0 & 89.8 & 259.0 & 181.9 & 40.2 & 296.8 & \textbf{163.7} \\
HMR2$^{\star}$ & 78.7 & 67.5 & 36.6 & 37.4 & 53.3 & 108.2 & 82.6 & 55.2 & 109.9 & 77.9 & 31.0 & 73.3 & \textbf{67.6} \\
GVHMR$^{\star}$ & 79.0 & 72.8 & 32.4 & 37.2 & 53.8 & 111.2 & 86.8 & 57.5 & 109.9 & 76.5 & 32.6 & 66.4 & \textbf{68.0} \\
WHAM$^{\star}$ & 75.2 & 101.1 & 48.2 & 46.3 & 61.5 & 132.7 & 129.1 & 86.2 & 175.5 & 88.5 & 37.3 & 119.9 & \textbf{91.8} \\
RoboPEPP$^{\star}$ & 78.2 & 93.9 & 39.2 & 42.0 & 58.7 & 109.6 & 109.3 & 75.6 & 146.7 & 83.7 & 34.7 & 91.2 & \textbf{80.2} \\
HoRoPose$^{\star}$ & 118.0 & 180.9 & 75.8 & 62.9 & 67.2 & 149.4 & 162.5 & 84.0 & 177.7 & 112.8 & 53.1 & 376.3 & \textbf{135.0} \\
RoboPose$^{\star}$ & 76.1 & 106.8 & 42.2 & 52.1 & 56.3 & 115.7 & 111.8 & 76.5 & 126.4 & 87.1 & 34.0 & 88.6 & \textbf{81.1} \\
\textbf{HumanoidPose (Ours)} & 54.7 & 57.5 & 22.6 & 30.5 & 48.5 & 84.0 & 69.4 & 47.3 & 76.3 & 59.3 & 20.2 & 55.8 & \textbf{52.2} \\
\bottomrule
\end{tabular}}
\vspace{2mm}
\end{table}

\begin{table}[!htbp]
\centering
\scriptsize
\setlength{\tabcolsep}{3pt}
\renewcommand{\arraystretch}{0.9}
\caption{Per-robot G-MPJPE (mm) on Humanoids-3D. Robots: H1=UnitreeH1, G1=G1, BT1=BoosterT1, BK1=BoosterK1, BHL=BerkeleyHumanoidLite, EPM=EngineAI\_PM01, FGR3=FourierGR3, FN1=FourierN1, KS45=KuavoS45, PND=PNDAdamLite, STD=StanfordToddy, TK=TienKung. The \emph{Mean} column is the robot-balanced average reported in the main table.}
\label{tab:perrobot_sim_g_mpjpe_mm}
\vspace{1mm}
\resizebox{\linewidth}{!}{
\begin{tabular}{l|cccccccccccc|c}
\toprule
Method & H1 & G1 & BT1 & BK1 & BHL & EPM & FGR3 & FN1 & KS45 & PND & STD & TK & Mean \\
\midrule
HMR2 + GMR & 1937 & 2074 & 2830 & 4871 & 5003 & 1316 & 700 & 2299 & 919 & 1671 & 10940 & 353 & \textbf{2909} \\
GVHMR + GMR & 1068 & 2053 & 2277 & 4171 & 4088 & 1420 & 676 & 2243 & 971 & 1796 & 10158 & 255 & \textbf{2598} \\
WHAM + GMR & 2427 & 2086 & 2990 & 4929 & 4534 & 1388 & 675 & 2344 & 933 & 1373 & 11386 & 707 & \textbf{2981} \\
HMR2$^{\star}$ & 572 & 235 & 214 & 227 & 398 & 273 & 311 & 233 & 387 & 476 & 281 & 235 & \textbf{320} \\
GVHMR$^{\star}$ & 515 & 292 & 225 & 317 & 399 & 283 & 325 & 263 & 373 & 416 & 328 & 241 & \textbf{332} \\
WHAM$^{\star}$ & 609 & 699 & 813 & 657 & 778 & 1136 & 878 & 741 & 895 & 1053 & 555 & 824 & \textbf{803} \\
RoboPEPP$^{\star}$ & 1581 & 1762 & 602 & 1019 & 525 & 2989 & 1664 & 844 & 12450 & 1273 & 1340 & 3967 & \textbf{2501} \\
HoRoPose$^{\star}$ & 873 & 5982 & 398 & 483 & 404 & 723 & 1703 & 513 & 1270 & 1901 & 803 & 1106 & \textbf{1347} \\
RoboPose$^{\star}$ & 727 & 641 & 439 & 449 & 421 & 506 & 609 & 442 & 655 & 723 & 297 & 574 & \textbf{540} \\
\textbf{HumanoidPose (Ours)} & 157 & 176 & 144 & 216 & 174 & 215 & 211 & 173 & 198 & 223 & 146 & 165 & \textbf{183} \\
\bottomrule
\end{tabular}}
\vspace{2mm}
\end{table}

\begin{table}[!htbp]
\centering
\scriptsize
\setlength{\tabcolsep}{3pt}
\renewcommand{\arraystretch}{0.9}
\caption{Per-robot PCK@0.3 (\%) on Humanoids-3D. Robots: H1=UnitreeH1, G1=G1, BT1=BoosterT1, BK1=BoosterK1, BHL=BerkeleyHumanoidLite, EPM=EngineAI\_PM01, FGR3=FourierGR3, FN1=FourierN1, KS45=KuavoS45, PND=PNDAdamLite, STD=StanfordToddy, TK=TienKung. The \emph{Mean} column is the robot-balanced average reported in the main table.}
\label{tab:perrobot_sim_pck_0p3}
\vspace{1mm}
\resizebox{\linewidth}{!}{
\begin{tabular}{l|cccccccccccc|c}
\toprule
Method & H1 & G1 & BT1 & BK1 & BHL & EPM & FGR3 & FN1 & KS45 & PND & STD & TK & Mean \\
\midrule
HMR2 + GMR & 69.3 & 46.7 & 90.8 & 85.5 & 64.3 & 36.2 & 50.4 & 63.2 & 68.0 & 45.2 & 95.2 & 65.8 & \textbf{65.0} \\
GVHMR + GMR & 62.1 & 69.5 & 89.9 & 83.9 & 82.8 & 62.4 & 46.6 & 64.7 & 59.2 & 57.6 & 94.0 & 76.6 & \textbf{70.8} \\
WHAM + GMR & 53.2 & 38.3 & 86.7 & 77.6 & 63.1 & 43.3 & 63.9 & 72.3 & 53.2 & 48.9 & 95.1 & 51.3 & \textbf{62.2} \\
HMR2$^{\star}$ & 82.5 & 84.3 & 93.7 & 94.8 & 89.8 & 72.7 & 78.6 & 86.4 & 76.0 & 78.3 & 97.4 & 82.5 & \textbf{84.7} \\
GVHMR$^{\star}$ & 82.5 & 82.3 & 95.2 & 95.1 & 89.3 & 72.4 & 77.9 & 85.8 & 75.8 & 79.7 & 97.1 & 84.7 & \textbf{84.8} \\
WHAM$^{\star}$ & 83.1 & 71.4 & 90.1 & 91.9 & 87.5 & 64.2 & 63.8 & 77.9 & 64.5 & 73.5 & 94.8 & 71.9 & \textbf{77.9} \\
RoboPEPP$^{\star}$ & 82.6 & 74.8 & 93.2 & 93.4 & 87.7 & 66.2 & 72.9 & 81.3 & 68.2 & 75.9 & 94.8 & 78.8 & \textbf{80.8} \\
HoRoPose$^{\star}$ & 77.0 & 46.0 & 82.3 & 84.0 & 84.4 & 58.2 & 55.3 & 77.2 & 60.6 & 63.1 & 92.9 & 55.6 & \textbf{69.7} \\
RoboPose$^{\star}$ & 83.1 & 70.6 & 92.4 & 91.7 & 89.2 & 68.5 & 69.5 & 78.4 & 72.3 & 75.2 & 96.0 & 78.5 & \textbf{80.5} \\
\textbf{HumanoidPose (Ours)} & 90.2 & 87.4 & 97.5 & 96.6 & 92.1 & 81.7 & 82.4 & 87.9 & 84.9 & 87.9 & 98.8 & 89.6 & \textbf{89.8} \\
\bottomrule
\end{tabular}}
\vspace{2mm}
\end{table}

\begin{table}[!htbp]
\centering
\scriptsize
\setlength{\tabcolsep}{3pt}
\renewcommand{\arraystretch}{0.9}
\caption{Per-robot 2D MPJPE (px) on Humanoids-Real. Robots: H1=UnitreeH1, G1=G1, BT1=BoosterT1, BK1=BoosterK1, BHL=BerkeleyHumanoidLite, EPM=EngineAI\_PM01, FGR3=FourierGR3, FN1=FourierN1, KS45=KuavoS45, PND=PNDAdamLite, STD=StanfordToddy, TK=TienKung. DHRP-H1 is a single external benchmark and is not broken down by robot; see the main table. The \emph{Mean} column reproduces the main-table value.}
\label{tab:perrobot_real2d_2d_mpjpe_px}
\vspace{1mm}
\resizebox{\linewidth}{!}{
\begin{tabular}{l|cccccccccccc|c}
\toprule
Method & H1 & G1 & BT1 & BK1 & BHL & EPM & FGR3 & FN1 & KS45 & PND & STD & TK & Mean \\
\midrule
GVHMR + GMR & 114.8 & 57.7 & 56.6 & 81.3 & 133.7 & 40.6 & 35.3 & 55.9 & 152.0 & 87.2 & 82.9 & 47.8 & \textbf{78.8} \\
HMR2 + GMR & 375.8 & 284.1 & 275.2 & 198.1 & 233.3 & 239.9 & 312.4 & 148.5 & 198.2 & 229.5 & 330.7 & 147.8 & \textbf{247.8} \\
WHAM + GMR & 320.8 & 188.7 & 61.9 & 203.3 & 196.9 & 177.8 & 355.1 & 133.5 & 177.8 & 300.9 & 92.3 & 166.7 & \textbf{198.0} \\
GVHMR$^{\star}$ & 92.2 & 191.2 & 175.7 & 219.8 & 192.3 & 131.8 & 168.1 & 132.5 & 50.8 & 117.8 & 243.9 & 75.2 & \textbf{149.3} \\
HMR2$^{\star}$ & 140.3 & 162.5 & 165.3 & 213.4 & 196.3 & 127.9 & 112.1 & 119.4 & 39.1 & 279.8 & 338.0 & 61.7 & \textbf{163.0} \\
WHAM$^{\star}$ & 243.1 & 174.7 & 285.2 & 366.8 & 190.3 & 145.7 & 469.2 & 181.9 & 87.0 & 182.0 & 502.2 & 94.2 & \textbf{243.5} \\
RoboPEPP$^{\star}$ & 186.8 & 106.3 & 76.4 & 119.4 & 144.2 & 356.3 & 204.8 & 85.9 & 59.8 & 142.2 & 81.7 & 106.3 & \textbf{139.2} \\
HoRoPose$^{\star}$ & 270.6 & 124.0 & 93.2 & 63.6 & 128.7 & 97.5 & 154.4 & 121.2 & 75.8 & 192.1 & 90.4 & 128.5 & \textbf{128.3} \\
RoboPose$^{\star}$ & 138.5 & 293.4 & 91.7 & 191.7 & 228.3 & 137.7 & 211.4 & 113.7 & 69.0 & 146.9 & 105.1 & 97.8 & \textbf{152.1} \\
\textbf{HumanoidPose (Ours)} & 77.7 & 65.7 & 54.5 & 88.1 & 131.4 & 53.6 & 76.8 & 39.3 & 45.4 & 70.8 & 92.1 & 62.5 & \textbf{71.5} \\
\bottomrule
\end{tabular}}
\vspace{2mm}
\end{table}

\begin{table}[!htbp]
\centering
\scriptsize
\setlength{\tabcolsep}{3pt}
\renewcommand{\arraystretch}{0.9}
\caption{Per-robot NME on Humanoids-Real. Robots: H1=UnitreeH1, G1=G1, BT1=BoosterT1, BK1=BoosterK1, BHL=BerkeleyHumanoidLite, EPM=EngineAI\_PM01, FGR3=FourierGR3, FN1=FourierN1, KS45=KuavoS45, PND=PNDAdamLite, STD=StanfordToddy, TK=TienKung. DHRP-H1 is a single external benchmark and is not broken down by robot; see the main table. The \emph{Mean} column reproduces the main-table value.}
\label{tab:perrobot_real2d_nme}
\vspace{1mm}
\resizebox{\linewidth}{!}{
\begin{tabular}{l|cccccccccccc|c}
\toprule
Method & H1 & G1 & BT1 & BK1 & BHL & EPM & FGR3 & FN1 & KS45 & PND & STD & TK & Mean \\
\midrule
GVHMR + GMR & 0.1397 & 0.0779 & 0.1244 & 0.1067 & 0.1636 & 0.0624 & 0.0484 & 0.1031 & 0.3933 & 0.1100 & 0.1939 & 0.1222 & \textbf{0.1371} \\
HMR2 + GMR & 0.4595 & 0.3817 & 0.5025 & 0.2630 & 0.2941 & 0.3703 & 0.4279 & 0.2881 & 0.5128 & 0.2831 & 0.7731 & 0.3245 & \textbf{0.4067} \\
WHAM + GMR & 0.3851 & 0.2556 & 0.1309 & 0.2734 & 0.2404 & 0.2721 & 0.4906 & 0.2538 & 0.4600 & 0.3657 & 0.2154 & 0.3751 & \textbf{0.3098} \\
GVHMR$^{\star}$ & 0.1343 & 0.2613 & 0.3351 & 0.2905 & 0.2187 & 0.1967 & 0.2256 & 0.2362 & 0.1315 & 0.1411 & 0.5697 & 0.1772 & \textbf{0.2432} \\
HMR2$^{\star}$ & 0.1668 & 0.2128 & 0.2944 & 0.2837 & 0.2430 & 0.1943 & 0.1510 & 0.2184 & 0.1013 & 0.3233 & 0.7903 & 0.1487 & \textbf{0.2607} \\
WHAM$^{\star}$ & 0.3499 & 0.2354 & 0.5169 & 0.4876 & 0.2339 & 0.2061 & 0.6261 & 0.3644 & 0.2251 & 0.2117 & 1.1726 & 0.2274 & \textbf{0.4048} \\
RoboPEPP$^{\star}$ & 0.2219 & 0.1428 & 0.1545 & 0.1560 & 0.1737 & 0.5376 & 0.2749 & 0.1609 & 0.1548 & 0.1766 & 0.1907 & 0.2441 & \textbf{0.2157} \\
HoRoPose$^{\star}$ & 0.3366 & 0.1690 & 0.1982 & 0.0843 & 0.1485 & 0.1498 & 0.2086 & 0.2184 & 0.1962 & 0.2335 & 0.2105 & 0.2746 & \textbf{0.2023} \\
RoboPose$^{\star}$ & 0.1673 & 0.4010 & 0.1756 & 0.2514 & 0.2811 & 0.2106 & 0.2908 & 0.2111 & 0.1786 & 0.1827 & 0.2454 & 0.2314 & \textbf{0.2356} \\
\textbf{HumanoidPose (Ours)} & 0.0919 & 0.0888 & 0.1077 & 0.1187 & 0.1622 & 0.0847 & 0.1023 & 0.0799 & 0.1175 & 0.0892 & 0.2153 & 0.1574 & \textbf{0.1180} \\
\bottomrule
\end{tabular}}
\vspace{2mm}
\end{table}

\begin{table}[!htbp]
\centering
\scriptsize
\setlength{\tabcolsep}{3pt}
\renewcommand{\arraystretch}{0.9}
\caption{Per-robot PCK@0.1 (\%) on Humanoids-Real. Robots: H1=UnitreeH1, G1=G1, BT1=BoosterT1, BK1=BoosterK1, BHL=BerkeleyHumanoidLite, EPM=EngineAI\_PM01, FGR3=FourierGR3, FN1=FourierN1, KS45=KuavoS45, PND=PNDAdamLite, STD=StanfordToddy, TK=TienKung. DHRP-H1 is a single external benchmark and is not broken down by robot; see the main table. The \emph{Mean} column reproduces the main-table value.}
\label{tab:perrobot_real2d_pck0p10}
\vspace{1mm}
\resizebox{\linewidth}{!}{
\begin{tabular}{l|cccccccccccc|c}
\toprule
Method & H1 & G1 & BT1 & BK1 & BHL & EPM & FGR3 & FN1 & KS45 & PND & STD & TK & Mean \\
\midrule
GVHMR + GMR & 53.05 & 71.64 & 59.55 & 48.36 & 31.04 & 84.12 & 96.22 & 50.27 & 5.24 & 51.66 & 22.93 & 65.19 & \textbf{53.27} \\
HMR2 + GMR & 4.36 & 5.16 & 0.61 & 11.27 & 9.52 & 5.84 & 2.52 & 11.39 & 0.00 & 11.22 & 0.00 & 7.32 & \textbf{5.77} \\
WHAM + GMR & 2.32 & 21.39 & 54.62 & 4.08 & 15.43 & 8.03 & 0.00 & 36.52 & 0.00 & 21.32 & 11.95 & 14.05 & \textbf{15.81} \\
GVHMR$^{\star}$ & 49.73 & 11.10 & 3.92 & 0.00 & 19.02 & 21.24 & 24.34 & 10.97 & 49.45 & 47.59 & 0.00 & 38.32 & \textbf{22.97} \\
HMR2$^{\star}$ & 20.48 & 8.64 & 0.52 & 0.04 & 2.59 & 6.44 & 35.36 & 7.79 & 53.07 & 11.00 & 0.00 & 45.38 & \textbf{15.94} \\
WHAM$^{\star}$ & 1.41 & 12.25 & 0.25 & 0.00 & 12.72 & 16.81 & 2.98 & 5.43 & 3.92 & 18.62 & 0.00 & 15.45 & \textbf{7.49} \\
RoboPEPP$^{\star}$ & 21.07 & 39.64 & 34.52 & 33.10 & 20.74 & 0.43 & 10.02 & 28.73 & 30.38 & 28.26 & 18.75 & 15.61 & \textbf{23.44} \\
HoRoPose$^{\star}$ & 2.71 & 29.60 & 34.74 & 74.93 & 29.86 & 47.80 & 27.44 & 10.91 & 37.71 & 17.21 & 17.65 & 14.12 & \textbf{28.72} \\
RoboPose$^{\star}$ & 34.13 & 14.66 & 26.69 & 14.55 & 9.39 & 27.66 & 9.63 & 17.58 & 27.41 & 30.65 & 15.81 & 19.52 & \textbf{20.64} \\
\textbf{HumanoidPose (Ours)} & 68.50 & 71.89 & 63.80 & 57.45 & 57.47 & 74.32 & 65.00 & 73.57 & 60.28 & 73.16 & 20.91 & 57.26 & \textbf{61.97} \\
\bottomrule
\end{tabular}}
\vspace{2mm}
\end{table}

\section{Humanoids Dataset Details}
\label{app:dataset_details}

This appendix provides the full statistics, simulation and rendering
configuration, and quality-control procedure behind the \textbf{Humanoids}
benchmark summarized in the main paper. The original dataset counts are reproduced
from the released manifest and audit reports.

\subsection{Dataset Statistics}

The original synthetic component is constructed from AMASS human motions and replayed
in two simulators, yielding 12 embodiments, 3,706 unique source motions, 6,885
clips, and 732,181 RGB frames. Table~\ref{tab:dataset-statistics} summarizes
the original dataset and Roban extension separately, Table~\ref{tab:dataset-robots} gives the original per-robot
coverage, and Table~\ref{tab:dataset-splits} reports the motion-held-out split.
The original synthetic clips are rendered at $640\times480$ with per-clip frame rate and a length
of 2--120 frames.

\begin{table}[h]
\centering
\small
\caption{Statistics of the \textbf{Humanoids} benchmark and Roban extension. The
original synthetic and real-world 2D counts are from the released manifest.
Roban adds synthetic clips and real videos with joint states and estimated
camera-relative root poses.}
\label{tab:dataset-statistics}
\setlength{\tabcolsep}{5pt}
\begin{tabular}{lrr}
\toprule
Component & Clips & Frames \\
\midrule
Synthetic --- Isaac Sim & 4,751 & 502,816 \\
Synthetic --- MuJoCo & 2,134 & 229,365 \\
Synthetic --- total & 6,885 & 732,181 \\
Real-world 2D --- original & 365 & 56,437 \\
Roban --- synthetic & 574 & 60,232 \\
Roban --- real 3D & 24 & 23,170 \\
All components & 7,848 & 872,020 \\
\bottomrule
\end{tabular}

\vspace{4pt}

\begin{tabular}{ll}
\toprule
Property & Value \\
\midrule
Original embodiments / with Roban & 12 / 13 \\
Unique source motions & 3,706 \\
Original synthetic duration & 6\,h\,46\,m \\
Original real-world duration & 31\,m\,30\,s \\
Roban synthetic duration & $\sim$33\,m\,30\,s \\
Roban real-world duration & $\sim$12\,m\,54\,s \\
Resolution & $640\times480$ \\
Frame rate & 30\,fps \\
Clip length & 2--120 frames \\
Source subsets (clips) & BMLrub 3,432 / CMU 2,227 / HDM05 1,226 \\
\bottomrule
\end{tabular}
\end{table}

\begin{table}[h]
\centering
\small
\caption{Per-robot coverage of the synthetic component. ``Isaac / MuJoCo''
reports the clip count rendered by each simulator. The 12 embodiments span
19--31 controllable degrees of freedom and are shared across train, validation,
and test; this is not an unseen-robot benchmark.}
\label{tab:dataset-robots}
\setlength{\tabcolsep}{4pt}
\begin{tabular}{lrrrrr}
\toprule
Robot & DoF & Bodies & Clips & Frames & Isaac / MuJoCo \\
\midrule
BerkeleyHumanoidLite & 22 & 26 & 566 & 60,930 & 385 / 181 \\
BoosterK1            & 22 & 23 & 561 & 59,349 & 386 / 175 \\
BoosterT1            & 23 & 24 & 569 & 60,125 & 405 / 164 \\
EngineAI\_PM01       & 24 & 29 & 522 & 55,299 & 357 / 165 \\
FourierGR3           & 31 & 39 & 560 & 59,563 & 394 / 166 \\
FourierN1            & 23 & 29 & 610 & 65,440 & 425 / 185 \\
G1                   & 29 & 28 & 582 & 62,081 & 407 / 175 \\
KuavoS45             & 28 & 29 & 611 & 65,063 & 412 / 199 \\
PNDAdamLite          & 25 & 26 & 558 & 59,255 & 390 / 168 \\
StanfordToddy        & 22 & 33 & 582 & 61,238 & 390 / 192 \\
TienKung             & 20 & 24 & 577 & 61,281 & 394 / 183 \\
UnitreeH1            & 19 & 20 & 587 & 62,557 & 406 / 181 \\
\midrule
Total                & -- & -- & 6,885 & 732,181 & 4,751 / 2,134 \\
\bottomrule
\end{tabular}
\end{table}

\begin{table}[h]
\centering
\small
\caption{Motion-held-out split of the synthetic component. All renderings of one
source motion, across robots, simulators, and scenes, are assigned to the same
split, so no source motion crosses a split boundary.}
\label{tab:dataset-splits}
\setlength{\tabcolsep}{6pt}
\begin{tabular}{lrrrr}
\toprule
Split & Clips & Frames & Duration & Source motions \\
\midrule
Train & 5,400 & 574,381 & 5\,h\,19\,m & 3,000 \\
Validation & 652 & 69,914 & 0\,h\,39\,m & 330 \\
Test & 833 & 87,886 & 0\,h\,49\,m & 376 \\
\midrule
Total & 6,885 & 732,181 & 6\,h\,46\,m & 3,706 \\
\bottomrule
\end{tabular}
\end{table}

\begin{table}[h]
\centering
\small
\caption{Rendering composition of the synthetic component. Isaac scenes use a
static camera within a clip that is re-randomized across clips together with
robot placement, distance, and lighting; MuJoCo clips use one of two orbit
camera policies. No accepted clip uses synthetic occluders.}
\label{tab:dataset-rendering}
\setlength{\tabcolsep}{6pt}
\begin{tabular}{llrr}
\toprule
Renderer & Scene / camera policy & Clips & Frames \\
\midrule
Isaac & \texttt{city\_anchor} (10 anchors) & 1,861 & 183,513 \\
Isaac & \texttt{isaac\_simple\_warehouse} & 835 & 92,310 \\
Isaac & \texttt{isaac\_outdoor\_rivermark\_safe} & 760 & 83,068 \\
Isaac & \texttt{brownstone\_origin\_v1} & 607 & 67,484 \\
Isaac & \texttt{residential\_lobby\_origin\_v1} & 455 & 50,763 \\
Isaac & \texttt{isaac\_office\_origin\_v1} & 233 & 25,678 \\
MuJoCo & \texttt{fixed\_orbit} & 1,733 & 185,295 \\
MuJoCo & \texttt{weak\_follow} & 401 & 44,070 \\
\bottomrule
\end{tabular}
\end{table}

\subsection{Motion Retargeting and Simulation}

Human motions from the HDM05, CMU, and BMLrub subsets of AMASS are retargeted
to each robot's native joint configuration with GMR, producing robot-native
joint angles, root pose, and FK-derived body positions. Each trajectory is then
replayed in Isaac Sim and MuJoCo, and the synchronized RGB frames, camera
parameters, joint states, root poses, and Canonical-18 3D keypoints are
recorded directly from the simulator state.

Rendering is randomized across clips. Isaac clips use a static camera within a
clip that is re-randomized across clips, together with robot placement, camera
distance, and lighting; the replay synchronizes simulation and capture with a
single physics step followed by a zero-delta-time capture. MuJoCo clips use
either a fixed-orbit or a weak-follow camera policy. No accepted clip uses
synthetic occluders. Table~\ref{tab:dataset-rendering} lists the scene and
camera composition, and the per-robot Isaac/MuJoCo split is given in
Table~\ref{tab:dataset-robots}.

\subsection{Quality Control Procedure}

Because all synthetic supervision is derived from simulator state, quality
control is defined on the state-to-observation mapping rather than on manual
annotation. We verify (i) structural validity (required fields, finite values,
tensor shapes), (ii) joint-limit legality, (iii) root-correction invariance
(joint angles, root XY, and root rotation are unchanged by grounding, and the
recorded Z correction matches the applied offset), (iv) FK consistency between
the released body geometry and the robot model, (v) image-space reprojection
consistency between the released 3D keypoints, camera, and 2D labels, and
(vi) ground non-penetration. Table~\ref{tab:dataset-qc-checks} reports the
maximum deviation of each check over the full release.

\begin{table}[h]
\centering
\small
\caption{Synthetic quality-control checks over the full release. ``Scope'' is
the audited population; every check reports the maximum observed deviation or
the number of failures.}
\label{tab:dataset-qc-checks}
\setlength{\tabcolsep}{5pt}
\begin{tabular}{p{0.38\linewidth}p{0.28\linewidth}p{0.26\linewidth}}
\toprule
Check & Scope & Result \\
\midrule
Required fields / finite / shapes & 6,885 clips & 0 failures \\
Joint-limit legality & 6,885 clips & 0 violations (max $4.78{\times}10^{-5}$ deg) \\
Root-correction invariance ($q$, root XY, rotation) & 6,885 clips & 0 altered \\
FK consistency & 20.1M body points & max $2.50{\times}10^{-4}$ mm \\
2D reprojection consistency & 13.0M valid keypoints & max $1.48{\times}10^{-2}$ px \\
Ground non-penetration & 6,885 clips & 0 violations (max $8.46{\times}10^{-5}$ mm) \\
Source teleport removal & 6,887 $\rightarrow$ 6,885 clips & 2 confirmed \\
Split integrity (motion leakage) & 3,706 source motions & 0 cross-split \\
\bottomrule
\end{tabular}
\end{table}

\begin{table}[h]
\centering
\small
\caption{Grounding-route summary over the pre-exclusion QC universe (6,887
clips; the released manifest excludes the two teleport clips). ``Fallback
ratio'' is the fraction of clips handled by a fallback route; ``root
deviation'' is the support-gap P95 of the route (PBHC preserves the original
root exactly). The released package uses the v3 support-segment fallback.}
\label{tab:dataset-qc-grounding}
\setlength{\tabcolsep}{5pt}
\begin{tabular}{lrrrr}
\toprule
Version / route & \#clips & Penetration & Fallback ratio & Root deviation \\
\midrule
PBHC (declared contacts) & 1,199 & 0 & 0.0\% & 0\,m \\
Constant fallback (v2) & 5,688 & 0 & 82.6\% & 8.89\,cm \\
Support-segment fallback (v3, released) & 5,688 & 0 & 82.6\% & 3.47\,cm \\
\bottomrule
\end{tabular}
\end{table}

\paragraph{Trajectory grounding.} Foot-ground consistency is handled by a
route-based grounding stage. Clips with reliable contact information use a
physics-based contact-aware route (PBHC) that preserves the source root; the
remaining clips use a support-segment median root-Z fallback, in which the root
height is set from the supporting segment while the source joint angles and
airborne semantics are preserved. Table~\ref{tab:dataset-qc-grounding} compares
the grounding routes and shows that the released v3 fallback reduces the
support-gap P95 from 8.89\,cm to 3.47\,cm relative to the constant v2 fallback,
while geometry non-penetration and root preservation are maintained.

\paragraph{Teleport and outlier removal.} Source trajectories are screened for
joint-velocity and root-position discontinuities. Two clips containing
confirmed source teleports are excluded from the released manifest
(6,887 $\rightarrow$ 6,885), and no further trajectory is altered by the
grounding stage.

\subsection{Roban Synthetic Evaluation}
\label{app:roban_synthetic}

Table~\ref{tab:roban_synthetic} reports results on the synthetic Roban test set.

\begin{table}[t]
\centering
\caption{Synthetic 3D evaluation on Roban. $^{\star}$ denotes humanoid-specific adaptation.}
\label{tab:roban_synthetic}
\small
\setlength{\tabcolsep}{5pt}
\begin{tabular}{lrrrrr}
\toprule
Method & MPJAE$\downarrow$ & MPJPE$\downarrow$ & PA-MPJPE$\downarrow$ & G-MPJPE$\downarrow$ & PCK@0.3$\uparrow$ \\
\midrule
HMR2+GMR & 32.62 & 83.91 & 81.70 & 1993.54 & 77.80 \\
GVHMR+GMR & 26.98 & 65.54 & 59.86 & 1314.99 & 86.56 \\
WHAM+GMR & 45.92 & 127.49 & 126.68 & 1032.92 & 73.00 \\
HMR2$^{\star}$ & 16.84 & 65.05 & 62.43 & 253.18 & 85.54 \\
GVHMR$^{\star}$ & 15.47 & 64.00 & 61.54 & 281.74 & 86.39 \\
WHAM$^{\star}$ & 15.08 & 64.11 & 61.82 & 296.61 & 86.93 \\
\midrule
RoboPose$^{\star}$ & 18.74 & 62.51 & 60.11 & 414.19 & 86.12 \\
HoRoPose$^{\star}$ & 23.87 & 90.29 & 83.65 & 791.38 & 77.91 \\
RoboPEPP$^{\star}$ & 19.84 & 71.44 & 68.19 & 15127.65 & 81.94 \\
\midrule
\textbf{HumanoidPose} & \textbf{13.00} & \textbf{49.20} & \textbf{47.24} & \textbf{135.33} & \textbf{91.10} \\
\bottomrule
\end{tabular}
\end{table}

\subsection{Roban Real-World 3D Evaluation}
\label{app:roban}

We teleoperate a physical Roban humanoid while recording RGB video with a
calibrated external camera and joint states through the robot SDK. Four
AprilTags are mounted on the front, back, left, and right sides of the rigid
torso. Detected tags and calibrated tag-to-root transformations provide
estimates of the camera-relative root pose. The 24 sequences contain 23,170
frames and include arm movements, dances, two walking gaits with varied
directions and trajectories, squatting, and combinations of these actions.
The 574 synthetic clips (60,232 frames) follow the same train, validation,
and test protocol as the original benchmark. Real videos are used for
evaluation without real-world fine-tuning.

\section{Canonical-18 Keypoint Definition}
\label{app:canonical18}

To enable consistent pose annotation and evaluation, we define a
Canonical-18 keypoint representation for humanoid robots. The keypoints
describe the semantic full-body structure of a humanoid robot and are
instantiated according to each robot's kinematic structure and geometry.
Rather than strictly following URDF joint origins, each keypoint is defined
according to the semantic center of the corresponding body part. For rigid
body components, we prioritize stable geometric or semantic centers that
remain consistent across viewpoints.

\begin{figure}[h]
    \centering
    \includegraphics[width=0.7\linewidth]{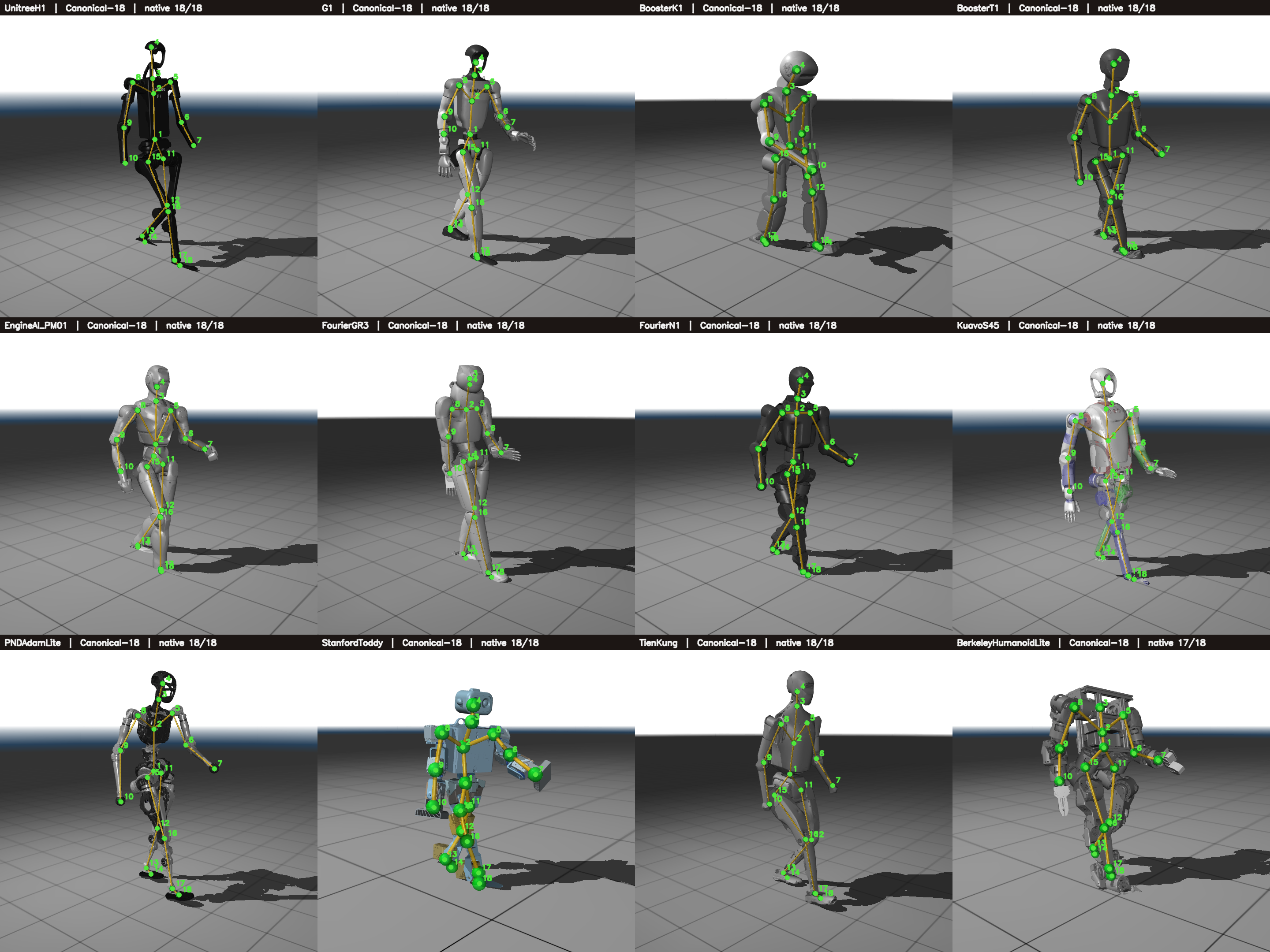}
    \caption{Visualization of the Canonical-18 keypoint representation for
    humanoid robots.}
    \label{fig:canonical18}
\end{figure}

Figure~\ref{fig:canonical18} illustrates the Canonical-18 keypoint layout.
The keypoints are designed to capture the major body parts of humanoid robots,
including the torso, head, upper limbs, and lower limbs.

The Canonical-18 keypoints are defined as follows:

\begin{table}[h]
\centering
\small
\begin{tabular}{c|l}
\toprule
Index & Keypoint \\
\midrule
1 & Pelvis \\
2 & Torso \\
3 & Neck \\
4 & Head \\
5 & L-Shoulder \\
6 & L-Elbow \\
7 & L-Wrist \\
8 & R-Shoulder \\
9 & R-Elbow \\
10 & R-Wrist \\
11 & L-Hip \\
12 & L-Knee \\
13 & L-Ankle \\
14 & L-Foot \\
15 & R-Hip \\
16 & R-Knee \\
17 & R-Ankle \\
18 & R-Foot \\
\bottomrule
\end{tabular}
\caption{Canonical-18 keypoint definition.}
\end{table}

For joint-like structures (e.g., shoulders, elbows, knees, and ankles),
keypoints are placed near the corresponding joint centers. For rigid
components (e.g., torso, head, and feet), keypoints represent the semantic
center of the visible structure rather than a specific mesh vertex or joint
origin. The left/right convention follows the robot's own perspective.

\section{Implementation Details of Kinematic Grounding and Refinement}
\label{app:kinematic_refinement}

\subsection{FK-conditioned Geometric Features}
\label{app:fk_features}

The proposed refinement module does not operate on the initial pose alone.
Instead, it receives explicit geometric features grounded by the robot
kinematic model.
Given the initial robot state $(q_0,R_0,t_0)$, we first obtain the metric
robot structure through forward kinematics:

\begin{equation}
    X_0=\operatorname{FK}_{r}(q_0),
\end{equation}

where $X_0\in\mathbb{R}^{J\times3}$ denotes the metric 3D robot structure
(Canonical-18 keypoints) in the root frame, determined by the
embodiment-specific kinematic tree and link geometry.

The initial 3D structure is transformed into the camera coordinate system:

\begin{equation}
    X_{0,j}^{C}=R_0X_{0,j}+t_0 .
\end{equation}

We then project the grounded structure into the image plane:

\begin{equation}
    \hat{x}_0=\Pi_K(X_0^C),
\end{equation}

and compute the reprojection residual:

\begin{equation}
    e_{\mathrm{reproj}}
    =
    \hat{x}-\hat{x}_0 .
\end{equation}

The final refinement feature is constructed by combining the initialized
state and FK-derived geometric cues:

\begin{equation}
\phi =
[
q_0,
R_0,
t_0,
(\hat{x},c),
\hat{x}_0,
e_{\mathrm{reproj}},
t_{\mathrm{direct}}-t_0
].
\end{equation}

Here $(\hat{x},c)$ contains observed 2D coordinates and confidence scores,
and $t_{\mathrm{direct}}$ is the direct translation estimate; its difference
from $t_0$ captures disagreement between translation initializers.
This explicit grounding converts the initial neural prediction into a
robot-specific geometric representation before refinement.

\subsection{Bounded Residual Parameterization}
\label{app:bounded_refinement}

The refinement module predicts bounded corrections instead of directly
regressing a new robot pose.
Specifically, the predicted residuals are parameterized as:

\begin{equation}
\begin{aligned}
    \Delta q &= \alpha_q\tanh(g_q),\\
    \Delta R &= \operatorname{Exp}([\alpha_R\tanh(g_R)]_{\times}),\\
    \Delta t &= \alpha_t\tanh(g_t),
\end{aligned}
\end{equation}

where $\alpha_q$, $\alpha_R$, and $\alpha_t$ control the maximum correction
magnitude.

The final robot state is updated as:

\begin{equation}
\begin{aligned}
    q
    &=
    \operatorname{clip}
    (q_0+\Delta q,q_{\min},q_{\max}),\\
    R
    &=
    \Delta R\,R_0,\\
    t
    &=
    t_0+\Delta t .
\end{aligned}
\end{equation}

The bounded formulation preserves the validity of the initial estimate and
prevents the refinement module from producing physically implausible states.
The joint-limit projection further guarantees that the refined articulation
remains within the feasible range of each robot embodiment.

\subsection{Training Objective}
\label{app:refinement_loss}

The refinement module is trained with multiple objectives defined in both
joint space and Cartesian space.
The overall objective is:

\begin{equation}
\mathcal{L}
=
\lambda_q\mathcal{L}_q
+
\lambda_{\mathrm{FK}}\mathcal{L}_{\mathrm{FK}}
+
\lambda_R\mathcal{L}_{R}
+
\lambda_t\mathcal{L}_{t}
+
\lambda_{\mathrm{cam}}\mathcal{L}_{\mathrm{cam}}
+
\lambda_{\mathrm{repr}}\mathcal{L}_{\mathrm{repr}}
+
\lambda_{\Delta}\mathcal{L}_{\Delta}.
\end{equation}

The joint-space loss is defined as:

\begin{equation}
\mathcal{L}_q
=
\operatorname{SmoothL1}(q,q^*).
\end{equation}

To explicitly enforce robot geometry consistency, we additionally apply a
forward-kinematic loss:

\begin{equation}
\mathcal{L}_{\mathrm{FK}}
=
\left\|
\operatorname{FK}_{r}(q)
-
\operatorname{FK}_{r}(q^*)
\right\|_1 .
\end{equation}

The root pose losses constrain the global camera-relative configuration, with
$d_{\mathrm{geo}}$ the geodesic distance on $\mathrm{SO}(3)$:

\begin{equation}
\mathcal{L}_{R}
=
d_{\mathrm{geo}}(R,R^*),
\end{equation}

\begin{equation}
\mathcal{L}_{t}
=
\operatorname{SmoothL1}(t,t^*).
\end{equation}

We further apply camera-space body alignment and image-space reprojection
constraints:

\begin{equation}
\mathcal{L}_{\mathrm{cam}}
=
\frac{1}{|\mathcal{B}|}\sum_{j\in\mathcal{B}}\left\|
RX_j(q)+t-
(R^*X_j(q^*)+t^*)
\right\|_2 ,
\end{equation}

Here $\mathcal{B}$ indexes the valid robot bodies used for camera-space alignment.

\begin{equation}
\mathcal{L}_{\mathrm{repr}}
=
\sum_j w_j
\left\|
\Pi_K(RX_j(q)+t)-\hat{x}_j
\right\|_2 .
\end{equation}

Finally, a residual regularization term encourages small articulation corrections:

\begin{equation}
\mathcal{L}_{\Delta}
=
\|\Delta q\|_2^2 .
\end{equation}

\subsection{Inference Behavior}
\label{app:refinement_inference}

The refinement module is a feed-forward prediction module.
During inference, it performs a single residual update from the initialized
state.
No test-time optimization, inverse kinematics fitting, or parameter update is
performed.

Therefore, the proposed refinement differs from optimization-based pose
fitting methods: the robot kinematic model is used as an explicit geometric
constraint during both training and inference, while the final correction is
predicted directly by the learned refinement network.

\section{Additional Implementation Details}
\label{app:implementation}

This section lists the concrete network sizes and training settings behind the
frozen-model numbers reported in the main paper. All components share the same
frozen visual encoder and are trained independently per robot.

\subsection{Network Architecture}

\paragraph{Visual encoder.}
We use a frozen V-JEPA ViT-B video encoder with patch size $16$, tubelet size
$2$, embedding dimension $768$, and $12$ transformer blocks, operating on
$16$-frame clips at $256\times256$ resolution. The input frames are the
detector crops, normalized using the ImageNet mean and standard deviation.
Each $16$-frame window is encoded once into $8$ tubelets of
$16\times16$ spatial tokens ($768$-d). The encoder contains $86.8$M parameters;
all of them are frozen (\texttt{requires\_grad=False}, eval mode) and only the
task heads are optimized.

\paragraph{Articulation head.}
The native-$q$ branch attends from $18$ learnable keypoint-conditioned queries ($256$-d) to
the target tubelet's spatial tokens (projected by LayerNorm and a
$768\!\to\!256$ linear layer); each query is conditioned on the corresponding
predicted Canonical-18 keypoint through an MLP $3\!\to\!256$, using its
two coordinates and confidence score. It consists of
one $8$-head cross-attention block ($256$-d, dropout $0.1$) followed by a
feed-forward block $256\!\to\!512\!\to\!256$ (GELU, dropout $0.1$), totaling
$\sim$$3.1$M trainable parameters.

\paragraph{Root pose heads.}
The root rotation is predicted by a two-hidden-layer MLP
($768{+}128{+}8 \to 512 \to 256 \to 6$) with LayerNorm, GELU and dropout
$0.1$, regressing a $6$D rotation ($\sim$$0.6$M parameters). The
camera-relative root translation is predicted by a separate simple head
($768{+}128{+}8 \to 512 \to 512 \to 3$, LayerNorm and GELU) whose three outputs
are converted analytically into a full-image camera translation
($\sim$$0.7$M parameters). Both heads consume the frozen mean-pooled target
tubelet token, the predicted Canonical-18, the normalized bounding box and the
normalized camera intrinsics.

\paragraph{Refinement module.}
The bounded refiner is a three-layer MLP
(LayerNorm $\to$ Linear $d_{\mathrm{in}}\!\to\!384 \to 256 \to (d_q{+}6)$,
GELU) whose last layer is zero-initialized, with
$d_{\mathrm{in}} = d_q + 6 + 3 + 3{\cdot}18 + 2{\cdot}18 + 2{\cdot}18 + 3$
($\sim$$0.17$M parameters).

\subsection{Training Configuration}

\paragraph{Inputs and windows.}
All branches share the same data interface: $16$-frame clips sampled at
$4$\,FPS from $30$\,FPS source video, $256\times256$ detector crops, the
predicted Canonical-18 keypoints, the normalized bounding box, and calibrated
intrinsics. No additional photometric or geometric augmentation is applied
beyond the detector crops and the ImageNet normalization; clips are shuffled
within each epoch.

\paragraph{Optimization.}
All heads are optimized with AdamW. The articulation head is trained in mixed
precision (AMP) with gradient scaling, and the refinement module is trained on
cached frozen features.

\begin{table}[h]
\centering
\small
\begin{tabular}{l c c c c}
\toprule
Component & LR & Weight decay & Batch size & Epochs \\
\midrule
Articulation head ($q$)   & $1\mathrm{e}{-4}$ & $1\mathrm{e}{-3}$ & 8   & 15 \\
Root rotation head        & $1\mathrm{e}{-4}$ & $1\mathrm{e}{-3}$ & 16  & 15 \\
Root translation head     & $1\mathrm{e}{-4}$ & $1\mathrm{e}{-4}$ & 8   & 15 \\
Refinement module & $2\mathrm{e}{-4}$ & $1\mathrm{e}{-3}$ & 128 & 15 \\
\bottomrule
\end{tabular}
\caption{Optimization settings. All runs use AdamW; the visual encoder is
frozen and receives no gradient. Model selection uses the validation split
only.}
\label{tab:impl_optim}
\end{table}

\paragraph{Objectives.}
The articulation head minimizes
$\mathcal{L}_q + 0.1\,\mathcal{L}_{\mathrm{FK}}$ with
$\mathcal{L}_q=\operatorname{SmoothL1}(q,q^*)$ and
$\mathcal{L}_{\mathrm{FK}}=\|\operatorname{FK}_r(q)-\operatorname{FK}_r(q^*)\|_1$;
the per-joint smooth-L1 term is re-weighted by the physical joint range.
The root rotation head uses a geodesic loss $\mathcal{L}_R=d_{\mathrm{geo}}$
and the translation head a smooth-L1 loss with a normalized target. The
refinement module combines
$\mathcal{L}_q + 0.1\,\mathcal{L}_{\mathrm{FK}} + 10^{-3}\|\Delta q\|_2^2
+ 0.5\,\mathcal{L}_R + 0.5\,\mathcal{L}_t
+ 0.1\,\mathcal{L}_{\mathrm{cam}} + 0.1\,\mathcal{L}_{\mathrm{repr}}$,
with gradients clipped to norm $1.0$. Its residual bounds are
$\alpha_q = 0.1\,(q_{\max}-q_{\min})$, $\alpha_R = 10^\circ$ and
$\alpha_t = 0.2$\,m.

\paragraph{Computational cost.}
{ We use the same two-stage recipe for every robot. We report the number of
\emph{training windows} rather than source clips: for G1, $2{,}629$ valid
training windows are extracted from $334$ rendered training clips, and each window provides $8$ tubelet
supervision targets, i.e., $2{,}629\times8=21{,}032$ window--tubelet targets.
Across the original $12$ robots the same procedure yields $28{,}387$ training windows
from $3{,}655$ clips (between $1{,}683$ and $2{,}867$ windows per robot).} The
articulation head is trained for $15$ epochs over these windows and accounts
for almost all of the cost ($\sim$$2$\,h on a single RTX~4090); caching the
frozen features for the refinement stage takes $\sim$$5$\,min, and the
refinement module itself $\sim$$2$\,min. Training one embodiment therefore
takes about $2$--$3$\,h on one GPU.

\paragraph{Robot-specific training.}
Each robot is trained independently with its own \texttt{RobotSpec} (kinematic
tree, joint limits, and Canonical-18 binding). We report the macro average over
the $12$ humanoid embodiments used in the main experiments.

\section{Limitations}
\label{app:limitations}

The main benchmark evaluates a fixed set of $12$ embodiments, with an additional
real-world 3D evaluation on Roban. Each robot is trained separately, and
we do not study a single model that
generalizes across embodiments, and evaluation on unseen robot morphologies
(zero-shot cross-embodiment transfer) is left to future work. In addition, our
real-world 3D evaluation is limited to one physical Roban robot; the other
real-world datasets provide manually verified \emph{2D} keypoints rather than
metric 3D annotations. Finally, we assume a single humanoid in view and do not address
multi-robot or human-robot interaction scenes.

\section{Evaluation Metrics}
\label{sec:metrics}

We evaluate humanoid pose recovery from both joint-space and Cartesian-space
perspectives. For synthetic evaluation, where robot joint states and
FK-derived 3D keypoints are available, we report joint angle error, 3D
keypoint errors, and pose alignment metrics; these metrics also apply to the
Roban real-world 3D evaluation. For the other real-world datasets, we evaluate
image-space keypoint recovery using manually verified 2D annotations.

\subsection{Synthetic 3D Pose Metrics}

\paragraph{Mean Per-Joint Angle Error (MPJAE).}
MPJAE measures the average angular difference between the predicted and
ground-truth robot joint configurations:
{
\begin{equation}
\mathrm{MPJAE} =
\frac{1}{D_r}\sum_{j=1}^{D_r}
\left\|
q_j-\hat{q}_j
\right\|_2 ,
\end{equation}
}
where \(q_j\) and \(\hat{q}_j\) denote the ground-truth and predicted angle of
the \(j\)-th robot joint, and \(D_r\) is the number of robot joints (the
joint-space dimension defined in Sec.~\ref{sec:problem_formulation}).
{Note that \(D_r\), the number of joint angles, is in general different from
\(J\), the number of Canonical-18 keypoints used by the position metrics
below.}

\paragraph{Mean Per-Joint Position Error (MPJPE).}
MPJPE measures the Euclidean distance between predicted and ground-truth
3D keypoints:

\begin{equation}
   \mathrm{MPJPE} =
\frac{1}{J}\sum_{j=1}^{J}
\left\|
X_j-\hat{X}_j
\right\|_2 ,
\end{equation}

where \(X_j\) and \(\hat{X}_j\) denote the ground-truth and predicted 3D
keypoints.

\paragraph{Procrustes-Aligned MPJPE (PA-MPJPE).}
PA-MPJPE evaluates pose structure after removing global translation,
rotation, and scale differences through Procrustes alignment:

\begin{equation}
 \mathrm{PA\text{-}MPJPE}
=
\frac{1}{J}\sum_{j=1}^{J}
\left\|
X^{*}_j-\hat{X}^{*}_j
\right\|_2 ,
\end{equation}

where \(X^{*}\) and \(\hat{X}^{*}\) denote the aligned poses.

\paragraph{Global MPJPE (G-MPJPE).}
G-MPJPE evaluates the absolute 3D pose accuracy in the camera coordinate
system without alignment:

\begin{equation}
 \mathrm{G\text{-}MPJPE}
=
\frac{1}{J}\sum_{j=1}^{J}
\left\|
X^{C}_j-\hat{X}^{C}_j
\right\|_2 .
\end{equation}

\paragraph{Percentage of Correct Keypoints (PCK).}
{ PCK@$\alpha$ measures the percentage of root-relative 3D keypoints whose
Euclidean error is below a fraction $\alpha$ of the robot's torso length:
}
\begin{equation}
\mathrm{PCK}@\alpha =
\frac{1}{J}
\sum_{j=1}^{J}
\mathbb{I}
\left(
\left\|X_j-\hat{X}_j\right\|_2
<
\alpha\,L_{\mathrm{torso}}
\right),
\end{equation}
where $X_j$ and $\hat{X}_j$ are the ground-truth and predicted root-relative
Canonical-18 keypoints, respectively, $J$ is the number of keypoints, $L_{\mathrm{torso}}$
is the robot's torso length, and $\mathbb{I}(\cdot)$ is the indicator
function. {The threshold ratio $\alpha$ is dimensionless; for synthetic 3D
evaluation we use $\alpha=0.3$ (i.e., errors below $30\%$ of the torso
length). The resulting fraction is multiplied by 100 when reported as a percentage.}

\subsection{Real-world 2D Pose Metrics}

For the real-world 2D datasets, which do not provide metric 3D ground truth, we evaluate
image-space keypoint recovery using manually verified Canonical-18 2D
annotations.

\paragraph{2D Mean Per-Joint Position Error (2D MPJPE).}
2D MPJPE measures the average pixel distance between predicted and annotated
keypoints:

\begin{equation}
 \mathrm{2D\text{-}MPJPE}
=
\frac{1}{J}\sum_{j=1}^{J}
\left\|
x_j-\hat{x}_j
\right\|_2 .
\end{equation}

\paragraph{Normalized Mean Error (NME).}
NME normalizes the 2D keypoint error by the size of the target bounding box:

\begin{equation}
 \mathrm{NME}
=
\frac{
\frac{1}{J}\sum_{j=1}^{J}
\left\|
x_j-\hat{x}_j
\right\|_2
}
{d_{\mathrm{bbox}}},
\end{equation}

where \(d_{\mathrm{bbox}}=\max(w,h)\), with bounding-box width \(w\) and height \(h\).

\paragraph{2D Percentage of Correct Keypoints (PCK).}
PCK measures the ratio of keypoints within a normalized distance threshold:

\begin{equation}
 \mathrm{PCK}@\alpha =
\frac{1}{J}
\sum_{j=1}^{J}
\mathbb{I}
\left(
\frac{\|x_j-\hat{x}_j\|_2}{d_{\mathrm{bbox}}}<\alpha
\right).
\end{equation}

We report PCK@0.1 for real-world evaluation, multiplying the resulting fraction by 100 to express it as a percentage.

\section{Baseline Details}
\label{sec:baseline_details}

We evaluate representative \emph{Human Pose Estimation (HPE)} and \emph{Robot
Pose Estimation (RPE)} methods under a unified HRPE protocol. Every baseline
output is converted into the robot-native joint configuration $q$, with 3D
keypoints obtained through the target embodiment's forward kinematics, and all
methods are scored with the same evaluator, target localization, and data splits
as \textbf{HumanoidPose}.

\paragraph{HPE baselines.}
We consider HMR2~\citep{goel2023humans}, GVHMR~\citep{shen2024world}, and
WHAM~\citep{shin2024wham}, which produce human-centric whole-body pose
representations. We evaluate two settings.
(1)~\emph{Direct application with human-to-robot retargeting}: no additional
training is performed; the original HPE weights, tracker, and keypoint estimator
are preserved, and only the input crop is replaced by the shared frozen robot
bounding box used for all methods (see the shared evaluation protocol below).
The predicted SMPL(-X) bodies are mapped to the target humanoid's native joints
with a frozen GMR retargeter~\citep{araujo2025retargeting}, followed by target
localization and camera-relative root composition.
(2)~\emph{Robot-specific adaptation}: the output representation is replaced by a
robot-specific head that predicts the native joint configuration and the
camera-relative root pose, trained on the Humanoids synthetic split while the
visual frontend is kept frozen.

\paragraph{RPE baselines.}
We consider RoboPose~\citep{labbe2021single}, HoRoPose~\citep{ban2024real}, and
RoboPEPP~\citep{goswami2025robopepp}. We adapt their output representation to each
target humanoid embodiment and train them on the corresponding synthetic data
following their original formulations. As these methods target seen embodiments,
they are reported only under the adaptation setting.

\paragraph{Shared evaluation protocol.}
For all methods, the robot is defined by its Canonical-18 keypoint specification
and kinematic model, and forward kinematics is computed with the same
implementation. Target localization is shared across methods: all baselines use
the same frozen robot bounding box for input cropping and the same normalization
scale $s=\max(w,h)$ for NME and PCK, so that per-method detectors do not affect
the main ranking. For real-world 2D evaluation we assume a fixed intrinsic matrix
($f_x=f_y=\max(W,H)$, $c_x=W/2$, $c_y=H/2$), whereas 3D evaluation uses the
benchmark-provided calibrated intrinsics. On the internal real-world set, all
twelve embodiments use human-reviewed 2D annotations. The synthetic split is
motion-disjoint, the
internal real-world split is source-disjoint, and the external DHRP-H1 benchmark
is used only for zero-shot evaluation.

\paragraph{Reporting rules.}
We never rank zero-shot and adapted baselines together without stating the
setting, and we do not mix different crop sources, test subsets, or evaluators in
a single ranked table. Failed detections are retained in the denominator, and all
methods in a table share the same bounding box, normalization scale, split, and
evaluator version.

\paragraph{Fairness of comparison.}
All robot-adapted baselines and HumanoidPose are trained on the same synthetic
splits under the same per-embodiment protocol, while directly transferred HPE
baselines are evaluated without additional training, and are evaluated with the same bounding boxes, normalization, splits,
and evaluator. Target localization and evaluation are never tuned per method, and
model selection uses the same held-out validation split. The HPE and RPE variants
therefore differ from \textbf{HumanoidPose} primarily in their pose
representation and prediction module, isolating the effect of the proposed
robot-native design.

\section{Real-2D PCK Threshold Sweep}

Tables~\ref{tab:real2d_pck_sweep_internal} and~\ref{tab:real2d_pck_sweep_dhrp}
report PCK at thresholds 0.3, 0.1, and 0.05 under exactly the same contract
as Table~\ref{tab:real2d_pose_estimation}: the Humanoids-Real column is the
12-robot robot-balanced average under the shared frozen robot bounding box, and
the DHRP-H1 column uses the external instance-id contract. PCK@0.1 is the
primary threshold reported in the main paper.

\begin{table}[H]
\centering
\scriptsize
\setlength{\tabcolsep}{3pt}
\renewcommand{\arraystretch}{0.95}
\resizebox{\textwidth}{!}{
\begin{tabular}{c|l|ccccc}
\toprule
Category & Method & 2D MPJPE$\downarrow$ & NME$\downarrow$ & PCK@0.3$\uparrow$ & PCK@0.1$\uparrow$ & PCK@0.05$\uparrow$ \\
\midrule
\multirow{3}{*}{HPE} & HMR2 + GMR & 247.77 & 0.4067 & 38.98 & 5.77 & 1.61 \\
 & GVHMR + GMR & 78.82 & 0.1371 & 89.49 & 53.27 & 25.83 \\
 & WHAM + GMR & 197.97 & 0.3098 & 56.55 & 15.81 & 4.90 \\
\midrule
\multirow{3}{*}{HPE} & HMR2$^{\star}$ & 162.98 & 0.2607 & 73.04 & 15.94 & 4.46 \\
 & GVHMR$^{\star}$ & 149.27 & 0.2432 & 69.93 & 22.97 & 7.07 \\
 & WHAM$^{\star}$ & 243.53 & 0.4048 & 49.17 & 7.49 & 1.90 \\
\midrule
\multirow{3}{*}{RPE} & RoboPEPP$^{\star}$ & 139.18 & 0.2157 & 78.52 & 23.44 & 6.66 \\
 & HoRoPose$^{\star}$ & 128.32 & 0.2023 & 78.09 & 28.72 & 12.06 \\
 & RoboPose$^{\star}$ & 152.10 & 0.2356 & 74.38 & 20.64 & 6.57 \\
\midrule
\multirow{1}{*}{HRPE} & \textbf{HumanoidPose (Ours)} & 71.49 & 0.1180 & 90.60 & 61.97 & 42.13 \\
\bottomrule
\end{tabular}}
\vspace{-2mm}
\caption{PCK threshold sweep on Humanoids-Real (12 robots) under the shared evaluation protocol of Table~\ref{tab:real2d_pose_estimation}. All methods use the same frozen robot bounding box for NME/PCK normalization; 2D MPJPE is in pixels and all PCK values are percentages. PCK@0.1 is the primary threshold.}
\vspace{-2mm}
\label{tab:real2d_pck_sweep_internal}
\end{table}

\begin{table}[H]
\centering
\scriptsize
\setlength{\tabcolsep}{3pt}
\renewcommand{\arraystretch}{0.95}
\resizebox{\textwidth}{!}{
\begin{tabular}{c|l|ccccc}
\toprule
Category & Method & 2D MPJPE$\downarrow$ & NME$\downarrow$ & PCK@0.3$\uparrow$ & PCK@0.1$\uparrow$ & PCK@0.05$\uparrow$ \\
\midrule
\multirow{3}{*}{HPE} & HMR2 + GMR & 253.68 & 0.3285 & 49.47 & 10.61 & 2.29 \\
 & GVHMR + GMR & 97.62 & 0.1470 & 87.40 & 62.06 & 28.09 \\
 & WHAM + GMR & 491.47 & 0.6015 & 8.70 & 0.99 & 0.31 \\
\midrule
\multirow{3}{*}{HPE} & HMR2$^{\star}$ & 146.48 & 0.1941 & 87.48 & 15.73 & 1.91 \\
 & GVHMR$^{\star}$ & 98.81 & 0.1481 & 86.87 & 60.69 & 32.60 \\
 & WHAM$^{\star}$ & 406.34 & 0.5003 & 16.26 & 2.06 & 0.46 \\
\midrule
\multirow{3}{*}{RPE} & RoboPEPP$^{\star}$ & 282.25 & 0.3684 & 50.99 & 8.40 & 1.53 \\
 & HoRoPose$^{\star}$ & 208.18 & 0.2830 & 66.18 & 17.48 & 5.34 \\
 & RoboPose$^{\star}$ & 212.69 & 0.2901 & 66.03 & 14.73 & 4.05 \\
\midrule
\multirow{1}{*}{HRPE} & \textbf{HumanoidPose (Ours)} & 87.80 & 0.1363 & 86.41 & 74.20 & 49.54 \\
\bottomrule
\end{tabular}}
\vspace{-2mm}
\caption{PCK threshold sweep on DHRP-H1 under the same external contract as Table~\ref{tab:real2d_pose_estimation}. 2D MPJPE is in pixels and all PCK values are percentages. PCK@0.1 is the primary threshold.}
\vspace{-2mm}
\label{tab:real2d_pck_sweep_dhrp}
\end{table}

\end{document}